\documentclass[10pt]{article}

\usepackage{fullpage}
\usepackage{setspace}
\usepackage{parskip}
\usepackage{titlesec}
\usepackage[section]{placeins}
\usepackage{breakcites}
\usepackage{lineno}
\usepackage{hyphenat}
\usepackage{amsmath,amsfonts}
\usepackage{algorithmic}
\usepackage{array}
\usepackage[caption=false,font=normalsize,labelfont=sf,textfont=sf]{subfig}
\usepackage{textcomp}
\usepackage{stfloats}
\usepackage{url}
\usepackage{verbatim}
\usepackage{graphicx}
\usepackage{booktabs}
\usepackage{color}
\usepackage{soul}
\usepackage[table]{xcolor}
\usepackage{pifont}
\newcommand{\cmark}{\ding{51}}
\newcommand{\xmark}{\ding{55}}
\usepackage{makecell}

\def \ie {\emph{i.e.}~}
\def \eg {\emph{e.g.}~}

\PassOptionsToPackage{hyphens}{url}
\usepackage[colorlinks = true,
            linkcolor = black,
            urlcolor  = black,
            citecolor = black,
            anchorcolor = black]{hyperref}
\usepackage{etoolbox}

\usepackage{natbib}

\renewenvironment{abstract}
  {{\bfseries\noindent{\abstractname}\par\nobreak}\footnotesize}
  {\bigskip}

\titlespacing{\section}{0pt}{*3}{*1}
\titlespacing{\subsection}{0pt}{*2}{*0.5}
\titlespacing{\subsubsection}{0pt}{*1.5}{0pt}

\usepackage{authblk}

\usepackage{amsmath}
\usepackage{graphicx}
\usepackage[space]{grffile}
\usepackage{latexsym}
\usepackage{textcomp}
\usepackage{longtable}
\usepackage{tabulary}
\usepackage{booktabs,array,multirow}
\usepackage{amsfonts,amsmath,amssymb}
\providecommand\citet{\cite}
\providecommand\citep{\cite}

\newif\iflatexml\latexmlfalse

\AtBeginDocument{\DeclareGraphicsExtensions{.pdf,.PDF,.eps,.EPS,.png,.PNG,.tif,.TIF,.jpg,.JPG,.jpeg,.JPEG}}

\usepackage[utf8]{inputenc}
\usepackage[english]{babel}

\usepackage{float}

\begin{document}

\title{AdaTreeFormer: Few Shot Domain Adaptation for Tree Counting from a Single High-Resolution Image}

\author[1]{Hamed Amini Amirkolaee}
\author[1,2]{Miaojing Shi\thanks{Corresponding author.
E-mail: \tt\small mshi@tongji.edu.cn.}}
\author[1]{Lianghua He}
\author[3]{Mark Mulligan}

\affil[1]{College of Electronic and Information Engineering, Tongji University, Shanghai, China}
\affil[2]{Shanghai Institute of Intelligent Science and Technology, Tongji University, China}
\affil[3]{Department of Geography, King’s College London, London, U.K.}

\vspace{-1em}

\begingroup
\let\center\flushleft
\let\endcenter\endflushleft
\maketitle
\endgroup

\selectlanguage{english}
\begin{abstract}
The process of estimating and counting tree density using only a single aerial or satellite image is a difficult task in the fields of photogrammetry and remote sensing. However, it plays a crucial role in the management of forests. The huge variety of trees in varied topography severely hinders tree counting models to perform well. The purpose of this paper is to propose a framework that is learnt from the source domain with sufficient labeled trees and is adapted to the target domain with only a limited number of labeled trees. 
Our method, termed as AdaTreeFormer, contains one shared encoder with a hierarchical feature extraction scheme to extract robust features from the source and target domains. It also consists of three subnets: two for extracting self-domain attention maps from source and target domains respectively and one for extracting cross-domain attention maps. For the latter, an attention-to-adapt mechanism is introduced to distill relevant information from different domains while generating tree density maps; a hierarchical cross-domain feature alignment scheme is proposed that progressively aligns the features from the source and target domains.
We also adopt adversarial learning into the framework to further reduce the gap between source and target domains. Our AdaTreeFormer is evaluated on six designed domain adaptation tasks using three tree counting datasets, \ie Jiangsu, Yosemite, and London. Experimental results show that AdaTreeFormer significantly surpasses the state of the art, \eg in the cross domain from the Yosemite to Jiangsu dataset, it achieves a reduction of 15.9 points in terms of the absolute counting errors and an increase of 10.8\% in the accuracy of the detected trees' locations. The codes and datasets are available at \emph{\color{magenta}{https://github.com/HAAClassic/AdaTreeFormer}}.


\end{abstract}

\sloppy

\section{Introduction}\label{sec:Introduction}
Tree counting {and mapping individual trees} plays an important role in various applications such as forest management (\cite{ong2021framework}), urban planning (\cite{eisenman2019urban}), environmental monitoring (\cite{johnson2018count}), and ecological balance maintaining (\cite{hennigar2008novel}). Counting the number of trees manually is time-consuming, costly, and often results in subjective estimates. Hence, there is a growing demand for the development of fast and accurate tree counting methods from remote sensing images using advanced computer vision algorithms (\cite{ammar2021deep}). However, accurately counting trees of different types, sizes, and shapes can be a difficult task due to factors such as varying topography, lighting conditions, and {shadows} (\cite{yao2021tree,liu2021deep}). These difficulties increase when the aim is to count trees solely from a single aerial image without a 3D digital surface model (DSM). DSM generation using light detection and ranging (LiDAR) or aerial images needs expensive equipment and lengthy data processing (\cite{ghanbari2021individual,liu2013extraction,pahlavani20173d}), hence making it valuable to develop an algorithm for tree counting without relying on DSM data.

Deep learning has revolutionized the fields of photogrammetry and remote sensing by automating analysis, improving accuracy, and advancing applications such as land cover segmentation and object detection (\cite{bigdeli2019deep}, \cite{zhu2017deep}, \cite{van2022dam}).
Despite the significant achievements of deep neural networks (DNNs), improving their performance generally depends on the availability of a lot of labeled training data, which requires a costly and laborious work for data curation (\cite{dos2019unsupervised}). This challenge intensifies when a DNN encounters multiple distinct domains, where each domain, for instance in tree counting, refers to a specific scene (urban, countryside, farmland), imagery type (aerial or satellite), with different levels of tree densities, shadows or overlapping among individual trees. Consequently, bridging the gap between source and target domains through knowledge transfer becomes essential.

This leads to a key issue in realistic application: the cross-domain problem, \ie training and test data come from different domains; also termed as the domain-adaptation problem. Previous works on tree counting are mostly implemented as a standard supervised learning problem (\cite{chen2022transformer,weinstein2019individual,ammar2021deep,weinstein2020neon,machefer2020mask}) which may suffer from overfitting to varying degrees due to the specific characteristics of the dataset. This is why domain adaptation for tree counting attracts our attention. Most existing studies on domain adaptation techniques in remote sensing focus on land cover and land use classification (\cite{liu2014domain,tuia2016domain}), hyperspectral image classification (\cite{deng2018active}), and scene classification tasks (\cite{song2019domain}). Many of them use large amounts of labeled target data to finetune/retrain pre-trained models (\cite{othman2017domain,song2019domain,deng2018active}), yet the high labelling cost has driven the development of techniques that leverage a large number of unlabeled target data instead. A prominent approach of such kind utilizes image translation networks, such as cycle-consistent generative adversarial network (CycleGAN), to translate source domain images to the target domain. These translated images are then used to train the network for the target task (\cite{song2019domain,wang2019learning, han2020focus}). It should be noting that the image translation models prioritize high-level features and style transfer (\cite{song2019domain,wang2019learning}), potentially sacrificing fine-grained details like individuals or small objects in the scene (\cite{pang2021image}). This loss of detail can greatly impact the tree counting accuracy. Furthermore, a significant domain gap between the source and target data limits the effectiveness of leveraging unlabeled target data for domain adaptation. This necessitates the development of a novel methodology capable of achieving high accuracy in the target domain with only a few labeled target data.

Few-shot domain adaptation requires models trained on a source domain with sufficient labeled data to adapt to a target domain with only a limited number of labeled data, which remains largely unexplored in remote sensing/computer vision. 
The memory-based models (\cite{vinyals2016matching,du2021hierarchical,wu2024domain}) and model-agnostic meta-learning (MAML)(\cite{liu2022generating,finn2017model})
are popular methods for this problem. The memory-based models use stored information from a source domain to adapt to a new target domain but often struggle with memory management (\cite{vinyals2016matching}). The MAML tackles this problem via an outer loop of model initialization on source domain data and an inner loop of model updating on the target domain data (\cite{finn2017model,liu2022generating}); it however struggles with significant domain shifts between source and target data. As far as we have explored, neural linear transform (NLT) (\cite{wang2021neuron}) and few-shot crowd counting (FSCC) (\cite{reddy2020few}) are the only methods available for few-shot domain adaptation in object counting. Their methods are also proved 
to be inadequate to capture the heterogeneity among domains (\cite{vettoruzzo2024advances}).

Overall, few-shot tree counting across diverse landscapes necessitates a more robust adaptation approach due to significant scene and tree count variations. To obtain accurate tree density maps in the target domain, samples from the source and target domains need to be mapped into a shared feature space to ensure that the projected features are both distinct and domain-invariant. To achieve this, we resort to the transformer model, a prominent deep learning architecture that has suitable potential for domain adaptation in classification and segmentation tasks (\cite{wang2021max, dosovitskiy2020image}).
Unlike {convolutional neural networks (CNNs)}, which operate on local areas of the image, the transformer models long-range dependencies among visual features throughout the entire image using a global self-attention mechanism. This is realized by dividing the image into non-overlapping patches of a set size and then combining them with positional embeddings through linear embedding (\cite{dosovitskiy2020image}). Nevertheless, how to leverage the ability of the transformer to extract robust cross-domain features for tree counting remains an open question.

To address the above question, we propose an attention-to-adapt mechanism for tree counting based on the transformer in a few-shot domain adaptation setting, which we have named as AdaTreeFormer. The proposed network consists of a transformer-based encoder to convert the input image into a latent representation and a transformer-based decoder with domain adaptation heads to produce the final tree density maps. The AdaTreeFormer has three subnets, namely source, source-target, and target subnets. The encoding parts of these subnets are shared and rely primarily on the self-attention mechanism to extract rich features, as do conventional transformer models. In contrast, we introduce an attention-to-adapt mechanism for decoding: each decoder in the source and target subnets uses a self-domain attention, while the source-target subnet has a cross-domain attention for domain information exchange. This design enables the model to efficiently encode intra-domain dependencies in source and target domains and inter-dependencies between source and target domains while generating tree density maps. Moreover, to ensure effective domain alignment among the output features of the three subnets, we introduce a hierarchical cross-domain feature alignment loss in which the generated self-domain attention maps from the source and target subnets should be close to the generated cross-domain attention maps from the source-target subnet. Last, we adopt an adversarial training approach to further decrease the domain gap and learn domain-invariant feature representations for generating consistently good tree density maps for target images.

Overall, we for the first time propose a few-shot domain adaptation framework for tree counting, incorporating new ideas by leveraging a transformer structure. In summary, we make the following contributions:

\begin{itemize}
\item We propose an end-to-end few-shot domain adaptation framework based on transformer architecture for tree counting from a single high-resolution aerial image. 
\item We propose an attention-to-adapt mechanism that enforces the network to extract relevant information from different domains. 
\item We introduce a hierarchical cross-domain feature alignment loss to help the network to align the extracted self- and cross-domain attention maps. 
\item We show that our framework generalizes and adapts to new domains with a much lower average error on the tree counting than state of the art models.

\end{itemize}

\section{Related Work}\label{RelatedWork}
In this section, we briefly review the relevant works in three subsections: supervised tree counting, cross-domain object counting, and few-shot learning.

\subsection{Supervised tree counting}\label{sec:SupervisedTreeCounting}
Tree counting in dense forest environments can be challenging due to the close proximity of trees, making it difficult to separate individual trees using traditional image analysis techniques. The recent achievements of DNNs in object recognition tasks (\cite{wang2019automatic,ren2015faster}) have motivated scientists to adapt these networks to automatic tree detection and counting. The networks proposed for tree counting can be mainly classified into two groups, namely detection-based and regression-based networks, specified below.  

\subsubsection{Detection-based methods}\label{sec:DetectionBasedMthods}
Detection-based methods use object detection algorithms to identify and count individual trees. These methods typically use models such as YOLO (\cite{redmon2016you}) and faster region convolutional neural network (R-CNN) (\cite{girshick2015fast}), which are pretrained on large image corpus annotated with bounding boxes; next, tree detection models are finetuned on tree-specific datasets using transfer learning (\cite{machefer2020mask,ammar2021deep,weinstein2019individual}). \cite{weinstein2020neon} have made available an open-access dataset containing tree crown information from several locations across the USA. They develop a delineation technique to generate the required training data for tree counting using LiDAR data.  \cite{ammar2021deep} conduct a comparative study of four object detection architectures - namely Faster R-CNN, YOLOv3, YOLOv4, and EfficientNet - for accurately counting and locating palm trees from aerial images. 
\cite{lassalle2022deep} present a hybrid approach combining a CNN and watershed segmentation to detect individual tree crowns from remotely sensed images.

\subsubsection{Regression-based methods}\label{sec:RegressionBasedMthods}
Instead of trying to localize individual trees by detection, regression-based methods focus on producing a continuous measure of tree density in an image (\cite{chen2022transformer,liu2021deep,yao2021tree}). These methods involve learning a model to produce a density map of the tree distribution in an image, which is then integrated to give an estimate of the total number of trees in the scene. These methods offer several advantages over detection-based methods, particularly in challenging scenarios involving high levels of occlusion and background clutter in densely forested environments. Typically, a Gaussian filter is used to create the density map by calculating a weighted sum of the pixel values around each labeled tree point. 
\cite{liu2021deep} introduce a pyramidal encoder-decoder network, aiming to handle variations in tree scale distribution by merging information across multiple decoder paths during inference. Similarly, \cite{osco2020convolutional} leverage a CNN to estimate the density map and examine the impact of near-infrared band usage on the prediction quality. \cite{yao2021tree} explore the performance of different networks such as visual geometry group network (VGGNet), AlexNet, and a two-column CNN built on the VGGNet and AlexNet backbones to estimate tree density map. \cite{chen2022transformer} have developed a method that incorporates both CNNs and transformer blocks for tree density map estimation. This architecture enables them to capture the global context while retaining the local information needed to represent tree distribution in complex scenes accurately. Finally, \cite{amini2023treeformer} propose a semi-supervised tree counting network. It focuses on increasing network performance by exploiting unlabeled images to reduce the need for labeled data. However, the performance of this network is reduced when training and test data belong to different domains.

\subsection{Cross-domain object counting}\label{sec:crossDomainObjectCounting} 

In addition to the above-mentioned supervised tree-counting methods,
\cite{zheng2020cross} proposes a cross-domain object detector for oil palm tree counting using different satellite images, which uses adversarial learning to exploit the transferability between the source and the target domains. 
Cross-domain adaptation has been the research subject for other counting tasks. \cite{machefer2020mask} have developed a method for counting plants using drone images, which involves deploying a Mask R-CNN model and implementing transfer learning to minimize the need for extensive training data. \cite{wang2019learning} compile a comprehensive synthetic crowd dataset to initially form a model capable of producing improved accuracy on real-world datasets through subsequent fine-tuning.
To achieve this, they leverage similarity embedding CycleGAN (SE-CycleGAN) as an unpaired image translation network (\cite{zhu2017unpaired}) to translate the synthetic domain into the real domain and then use the translated images to estimate crowd density maps. In the SE-CycleGAN (\cite{wang2019learning}), the structural similarity index measure (\cite{wang2004image}) is used as an additional cycle loss to improve the translation performance.
\cite{li2019coda} introduce a counting object via density adaptation (CODA) to tackle varying object scales and density distributions through adversarial training using multi-scale pyramid patches from both the source and target domains. They achieve consistent object counts across different scales by applying a ranking constraint across the pyramid levels. \cite{reddy2021adacrowd} propose an adaptation for crowd counting (Adacrowd) that employ a guiding network to extract batch normalization parameters from the unlabeled images of the target domain for crowd counting.
\cite{liu2022discovering} formulate mutual transformations between regression- and detection-based models as two scene-agnostic transformers. They introduce a self-supervised co-training scheme on the target to finetune the regression and detection models using generated pseudo labels, thereby iteratively enhancing the performance of both.
\cite{du2023domain} introduce a domain-general crowd counting (DGCC) method for training a model on a single source domain to generalize to unseen domains in crowd counting. They divide the source domain into sub-domains for meta-learning and refine the division dynamically. Additionally, they utilize domain-invariant and domain-specific crowd memory modules to disentangle image features.

\subsection{Few-shot Learning}\label{sec:FewShotLearning}
Few-shot learning refers to the capability of a model to perform well on previously unseen tasks when presented with only a small amount of training data. Several approaches have emerged over the years: early methodologies 
(\cite{bart2005cross,fink2004object,fei2006knowledge,zhang2022few,zhao2021domain}) use hand-crafted features {such as local scale-invariant features to transfer knowledge from shared features and contextual information. While more recent ones, such as the memory-based models (\cite{vinyals2016matching,du2021hierarchical,wu2024domain}), LSTM-based meta-learning (\cite{ravi2017optimization,yang2024few}), prototypical networks (\cite{santoro2016meta, huang2022few}), MAML (\cite{finn2017model}) are based on learned features that are automatically extracted from DNN models. \cite{vinyals2016matching} employ memory components within a neural network to acquire a shared representation from minimal data. \cite{santoro2016meta} introduce the prototypical networks that acquire a metric space where classification is achieved by calculating distances to prototype representations of individual classes. Additionally, \cite{ravi2017optimization} employ an LSTM-based meta-learner to acquire an update rule for training a neural network, facilitating improved adaptation to new tasks with limited data. \cite{finn2017model} utilize a model-agnostic meta-learning (MAML) that learns an optimal model parameter initialization, enabling improved generalization to similar tasks through meta-learning.}
\cite{santoro2016meta} highlight the augmented memory neural network for remembering previous tasks and use it to train a model for new tasks. In parallel, \cite{mishra2017simple} introduce a general meta-learning architecture that combines temporal convolutions for aggregating past information with soft attention for focusing on specific details. \cite{liu2022generating} propose sharp-MAML to enhance the performance in few-shot meta-learning by leveraging sharpness-aware minimization.

Although most of the above few-shot learning techniques are used in classification tasks, there is also some research on few-shot object counting, which is mainly used for crowd counting. \cite{reddy2020few} present a few-shot crowd counting (FSCC) that employ the MAML to tune the model parameters in unseen scenes using a few labeled images. In this method, a new domain means a different camera position, where the scene is fixed and the population in front of the camera is varied. Our purpose is to perform few-shot domain adaptation in tree counting, where both the scene (landscape) and the number of trees vary across images in the new domain. This significant domain shift makes it much more challenging. Recently, \mbox{\cite{wang2021neuron}} propose the neural linear transform (NLT) to tackle few-shot domain adaptation by learning the domain factors and biases, so as to adjust model parameters to function effectively in the new domain. They only apply a linear transformation on model parameters which however is inadequate to adapt with the drastic domain shift.

\medskip
In general, the benefit of conventional tree counting networks lies in their ability to produce accurate tree numbers in a specific domain. It is crucial to provide a domain adaptation method due to the infeasibility of providing training data that covers all different domains.

\begin{figure*}[!t]
\centering
\includegraphics[width=6.5in]{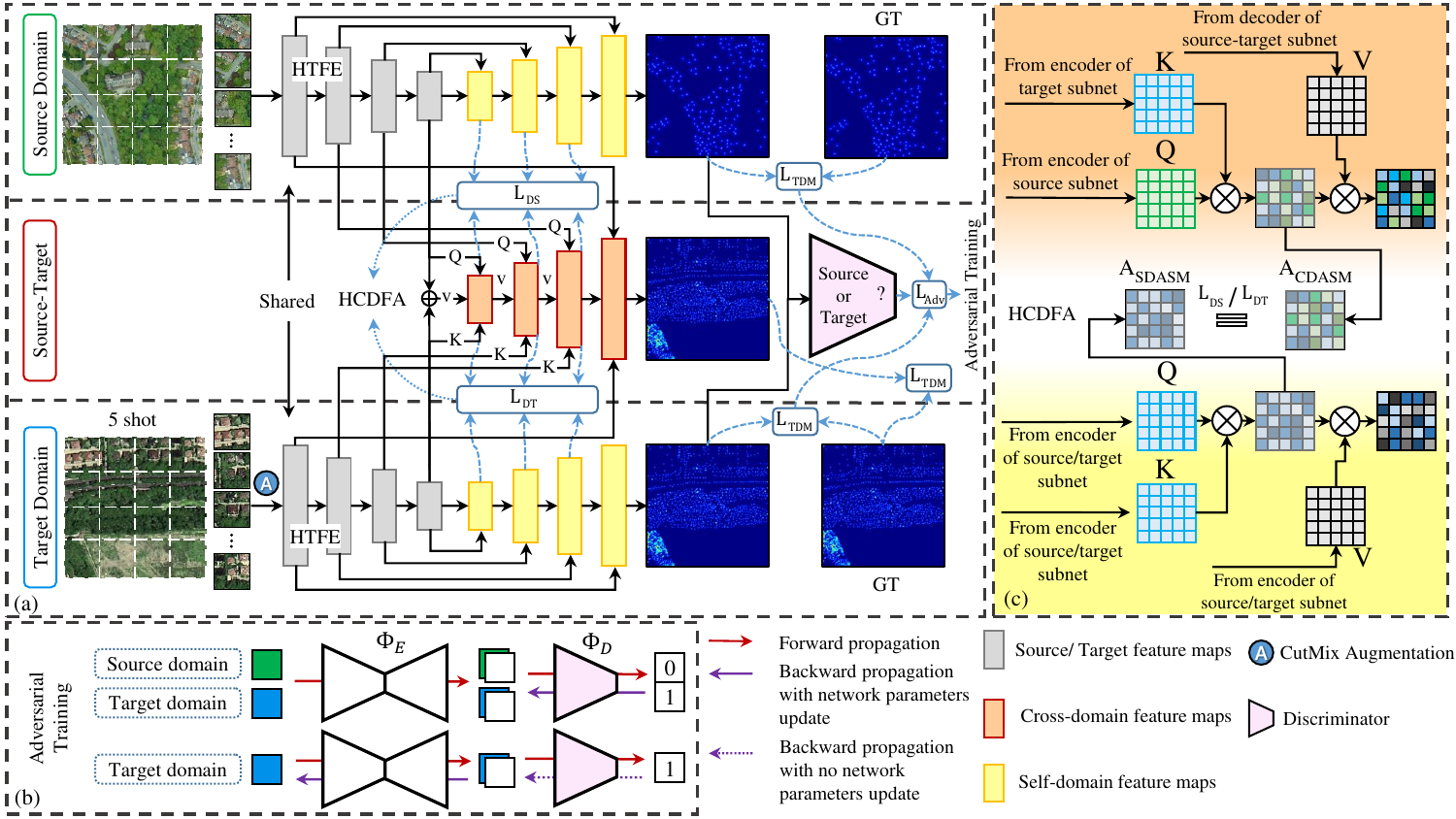}
\caption{Overview of the proposed AdaTreeFormer framework. (a) In the source domain subnet, the estimated tree density maps of the source domain are optimized with ground truth (GT) using $L_{TDM}$. The similar pipeline goes to the target domain subnet with the few-shot target domain images, obtained through Cutmix augmentation. The estimated tree density maps in the source-target subnet, produced after the cross-domain attention, are optimized using GT of the target domain through $L_{TDM}$, while the estimated feature maps are refined using $L_{DT}$ and $L_{DS}$. An adversarial training is employed that is optimized with $L_{Adv}$. (b) Explanation of the processes of advised adversarial training in one iteration. (c) The structure of the HCDFA using the self- and cross-domain attention score maps ($A_{SDASM}$ and $A_{CDASM}$) for a specific scale. {When the input feature maps in the lower yellow part are from the source subnet, the equality between $A_{SDASM}$ and $A_{CDASM}$ is equivalent to $L_{DS}$ in $L_{HCDFA}$, and when those feature maps are from the target subnet, this equality is equal to $L_{DT}$.}} 

\label{fig:overview}
\end{figure*}

\section{Methodology}\label{sec:Methodology}
\subsection{Overview}\label{sec:Overview}
This paper proposes a few-shot domain adaptation framework to estimate the density map of trees from a single remote sensing image. An overview of the designed framework is presented in Fig.~\ref{fig:overview}. Our framework has a hierarchical tree feature extraction (HTFE) module based on the transformer encoder (Section \ref{sec:HierarchicalVisionTransformer}) that has a shared weight for the source and target domains. 
Given an image from the source domain ($\mathcal{I}_s\in {\mathbb{R}^{H\times W \times 3}}$) and another one from the target domain ($\mathcal{I}_t\in {\mathbb{R}^{H\times W \times 3}}$), the HTFE generates multi-scale feature maps from them, $\mathcal{F}^s$ and $\mathcal{F}^t$, respectively. We design a domain adaptive decoder with three subnets: the source, target, and source-target subnets. 
The source subnet estimates the tree density map corresponding to $\mathcal{I}_s$ as $\mathcal{T}_{s}$ using $\mathcal{F}^s$, while the target subnet estimates the tree density map of $\mathcal{I}_t$ as $\mathcal{T}_{t}$ using $\mathcal{F}^t$. Additionally, the source-target subnet utilizes $\mathcal{F}^s$ and $\mathcal{F}^t$ to estimate the $\mathcal{T}_{st}$ corresponding to $\mathcal{I}_{t}$.
An attention-to-adapt mechanism is proposed to help the model learn domain-invariant representations from both domains. This mechanism plays the role of self-domain adaptation in the source and target subnets and cross-domain adaptation in the source-target subnet. By leveraging the self- and cross-domain attention maps, the model learns to adapt to the differences between domains and improve its performance on the target domain (Section \ref{sec:AttentiontoAdaptMechanism}).
A domain adaptive learning strategy is introduced that contains three parts including supervised learning, cross-domain learning, and adversarial learning. The tree distribution matching (TDM) loss is employed for the supervised learning of all subnets of the proposed framework (Section \ref{sec:TreeDistributionMatching}). We design the hierarchical cross-domain feature alignment (HCDFA) with the objective of aligning the self-domain attention maps with the cross-domain attention {score maps} (Section \ref{sec:HierarchicalCrossDomainFeatureAlignment}). Finally, a domain discriminator network (see pink trapezia in Fig. \ref{fig:overview}) is employed to determine the domain of the $\mathcal{T}_s$ and $\mathcal{T}_t$ (Section \ref{sec:DomainDiscriminator}).  We employ adversarial learning to make the feature distributions of the source and target subnets close (Section \ref{sec:AdversarialLearning}). {Overall, throughout the training process all subnets are utilized, while only the target subnet is employed for testing.}

\subsection{AdaTreeFormer Framework}\label{sec:AdaTreeFormerFramework}
This section includes two parts, one is the HTFE module that is based on the transformer as the encoder of our AdaTreeFormer. Moreover, we introduce the attention-to-adapt mechanism as the decoder of our model. They are also illustrated in Fig.~\ref{fig:Module_1} in detail.

\subsubsection{Hierarchical Tree Feature Extraction}\label{sec:HierarchicalVisionTransformer} 
\begin{figure*}[!t]
\centering
\includegraphics[width=6.5in]{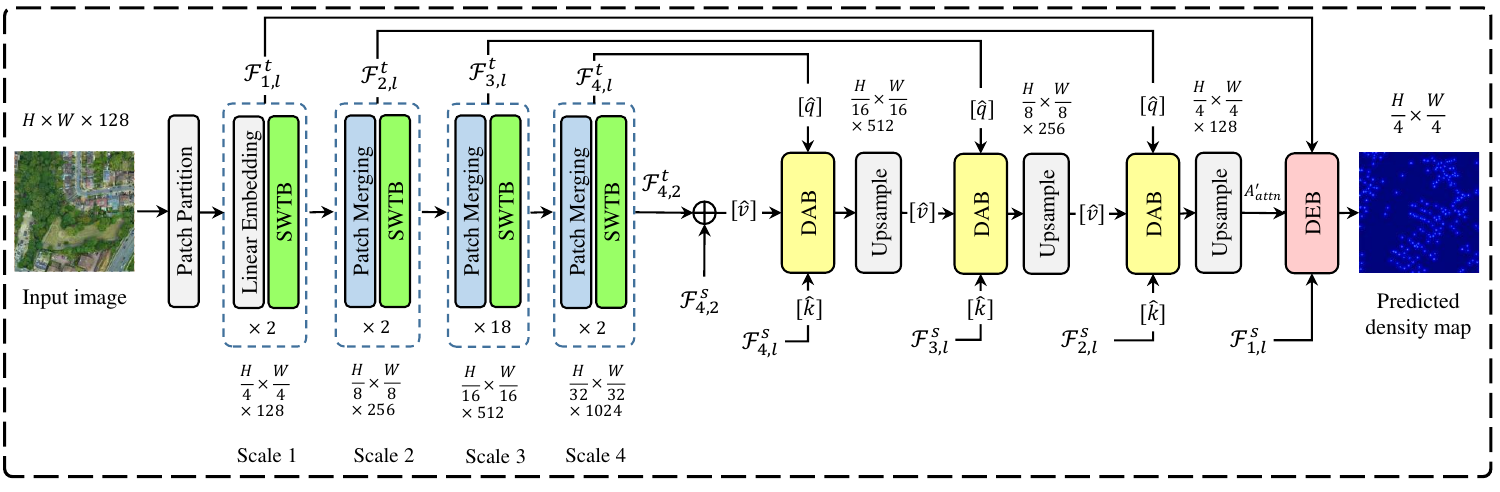}
\caption{(a) The details of the encoder-decoder part of the proposed AdaTreeFormer. Given the input image, multi-scale features are firstly extracted through the linear embedding, shifted windows transformer block (SWTB), and patch merging module in the encoder. The domain attention blocks (DAB) and density estimation block (DEB) in the decoder align the source and target domains and generate the tree density map, respectively.}

\label{fig:Module_1}
\end{figure*}

\begin{figure*}[!t]
\centering
\includegraphics[width=3in]{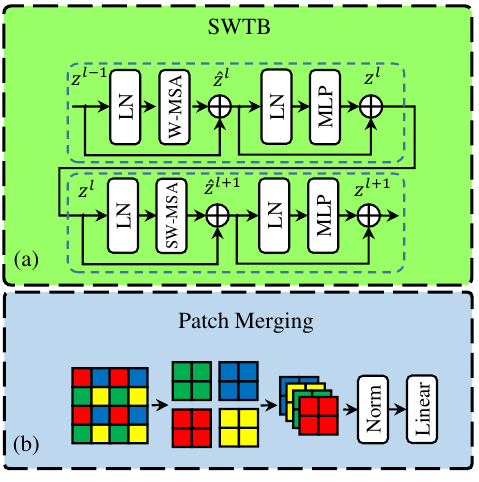}
\caption{(a) The structure of the SWTB for extracting feature maps. (b) The patch merging module incorporates information from different image patches into a single unified representation.}

\label{fig:Module_2}
\end{figure*}

We develop the HTFE based on the transformer (\cite{liu2021swin}) to effectively extract multi-scale features during the encoding part of the network (Fig.~\ref{fig:Module_1}). In the HTFE, the input image ($H\times W \times 3$) is divided into $4\times4$ non-overlapping patches, where each patch is considered a token and its feature is formed by concatenating the raw pixel RGB values. {Thus, the feature dimension of each patch is $4 \times 4 \times 3 = 48$.} In the first scale, a linear embedding layer projects each achieved patch into the dimension of $\frac{H}{4}\times \frac{W}{4}\times 128$. Then, the shifted window transformer block (SWTB) (Fig.~\ref{fig:Module_2}a) is applied to the patch (\cite{liu2021swin}).

The purpose of the shifted window transformer block (SWTB) (Fig.~\ref{fig:Module_2}a) is to efficiently capture long-range dependencies in an image by dividing it into shifted windows and applying self-attention within each window. This block is constructed by substituting the conventional multi-head self-attention (MSA) module in a transformer block with a module that relies on shifted windows. In the SWTB, the shifted window-based MSA module is preceded by a 2-layer multilayer perceptron (MLP) with GELU non-linearity in the middle. Each MSA module and MLP are preceded by a layer normalization (LN) and a residual connection is appended after each module, respectively (\cite{liu2021swin}).

In order to create a hierarchical multi-scale representation, the number of tokens is decreased by applying a patch merging layer (see Fig.~\ref{fig:Module_2}b) when the network gets deeper. 
The initial patch merging layer in the second scale concatenates the feature maps of each group of 2×2 neighboring patches and applies a linear layer to the concatenated feature maps, reducing the number of tokens by a factor of 4 ($2 \times 2$) and setting the output dimension to 256. Similarly, the patch merging and SWTB are applied in the third and fourth scales to generate feature maps with the resolution of $\frac{H}{16}\times \frac{W}{16}\times 512$ and $\frac{H}{32}\times \frac{W}{32}\times 1024$, respectively. Following (\cite{liu2021swin}), the encoder comprises 2 layers in the first, second, and fourth scales, while the third scale consists of 18 layers ($\tau \in \{2, 2, 18, 2\}$). $\tau$ represents the number of layers in each scale of the HTFE.

As seen in Fig.~\ref{fig:overview}a, {the HTFE is applied to both source and target data for extracting feature maps from them. The source domain contains a large number of labeled images, while} the target domain has only a limited number of labeled images available. Therefore, to strengthen the target training data and make the network unbiased, the Cutmix method (\cite{yun2019cutmix}) is used as a data augmentation technique to increase the training data of the target domain. 
While traditional methods apply transformations independently to individual images, CutMix combines two random images by cutting out a patch from one image and replacing it with a patch from the other. This creates a new training sample with mixed information that encourages the model to learn more robust and discriminative features by forcing it to rely on different parts of the two images. (\cite{yun2019cutmix}.

\subsubsection{Attention-to-Adapt Mechanism}\label{sec:AttentiontoAdaptMechanism}
The purpose of the attention-to-adapt mechanism is to help the model to focus on domain-invariant information from different domains while generating tree density maps. The mechanism is implemented across the three subnets of our framework. After passing the source and target images through the encoder, a collection of multi-scale feature maps are generated through HTFE, \ie $\mathcal{F}_{i,l}^s$ and $\mathcal{F}_{i,l}^t$, where $i$ is the scale of the feature maps with respect to the input image, $i \in \{1,2,3,4\}$; $l$ is the index of certain layer of a specific scale, $l \in \{1,..., \tau_i\}$; $s$ and $t$ indicate the source and target domains. These feature maps are considered as the input of our proposed attention-to-adapt mechanism. This mechanism comprises two major blocks including domain attention block (DAB) and density estimation block (DEB).

\begin{figure*}[!t]
\centering
\includegraphics[width=6.5in]{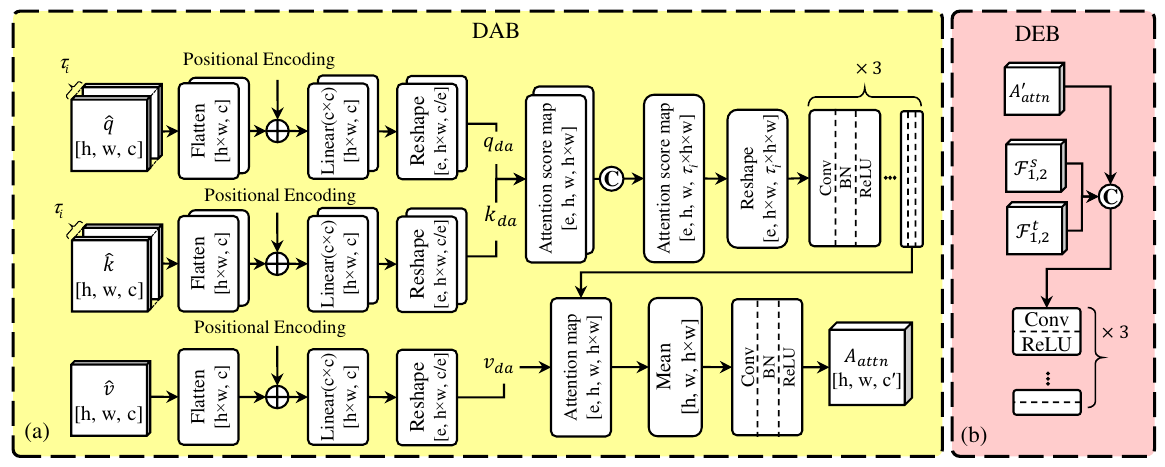}
\caption{(a) The structure of the DAB that generates the self- or cross-domain attention maps using the produced feature maps from the encoder part of the network, (b) The DEB estimates the final tree density map by fusing the generated feature maps from the encoder ($\mathcal{F}_{1,2}^s$ and $\mathcal{F}_{1,2}^t$) and the last layer of the decoder ($A_{attn}^{'}$).}

\label{fig:Module_3}
\end{figure*}

The DAB (Fig.~\ref{fig:Module_3}a) is to generate self-domain attention maps to ensure efficient encoding of intra-domain dependencies in both the source and target subnets. Additionally, it generates the cross-domain attention maps to account for the inter-dependencies between these domains in the source-target subnet. This block is based on the transformer (\cite{vaswani2017attention}) and employs positional encoding and multi-head attention to compute attention maps.

The DEB (Fig.~\ref{fig:Module_3}b) serves as the final block responsible for computing the tree density map (Fig.~\ref{fig:Module_1}). It achieves this by applying convolutional and Relu layers to the concatenation of $A_{attn}^{'}$ obtained from the preceding layer and $\mathcal{F}_{1,2}^s$ and $\mathcal{F}_{1,2}^t$ from the encoder, employing skip connections.

\textbf{Domain Attention Block:}
The DAB is designed to produce the self-/cross-domain attention maps using the generated feature maps through the encoder part of the network. Following the self-attention as the main component of the transformer architecture (\cite{vaswani2017attention}), the attention maps are computed as follows:

\begin{equation}\label{eq:Eq1}
    \ A_{attn}(q_{da}, k_{da}, v_{da}) = softmax(\frac{q_{da}k_{da}^T}{\sqrt{d}})v_{da}
\end{equation}

where $q_{da}$, $k_{da}$, and $v_{da}$ represent sets of query, key, and value vectors and $d$ denotes the dimension of the query and key vectors.
We adopt this equation to compute dense pixel-wise attention across extracted feature maps from the same or different domains. Therefore, the extracted feature maps from the encoder are specified as $\hat{q}$, $\hat{k}$, and $\hat{v}$. $\hat{q}$ can be from the same/different domains of $\hat{k}$ and $\hat{v}$. These matrices change depending on whether the aim is to extract self- or cross-domain attention maps. 
The HTFE generates $\tau_i$ layer feature maps for $i^{th}$ scale of the encoder (Section \ref{sec:HierarchicalVisionTransformer}). Hence, the $\hat{q}$ and $\hat{k}$ are multi-layer feature maps according to the $\tau_i$, while $\hat{v}$ initially contains the last-layer feature maps of the fourth scale of the encoder (see Fig. \ref{fig:Module_3}a). The dimension of $\hat{v}$ and each layer of $\hat{q}$, $\hat{k}$, is equal to $[h\times w\times c]$, where $h$, $w$, and $c$ are the height, width, and channel number, respectively. We flatten the $\hat{v}$ and each layer of $\hat{q}$ and $\hat{k}$ into the dimension of $[h\times w, c]$ to treat each pixel as a token and a positional encoding is added to it. Similar to the original transformer (\cite{vaswani2017attention}), we employ sine and cosine functions with different frequencies for positional encoding and utilize multi-head attention. After adding the positional encoding, a linear projection  $(c\times c)$ is applied, and the outcomes of this projection are reshaped according to the number of heads ($e$) of the multi-head attention to achieve the $q_{da}$, $k_{da}$, and $v_{da}$. The attention score maps are computed using $softmax(q_{da}k_{da}^T / \sqrt{d})$, resulting in a dimension of $[e, h, w, h\times w]$. The resulting multi-layer attention {score maps} are concatenated together, obtaining attention {score maps} with the dimension of $[e, h, w, \tau_i \times h\times w]$ (see Fig. \ref{fig:Module_3}a). After reshaping the concatenated maps, they are passed through three consecutive series of convolutional, batch normalization, and ReLU layers, and the resulting computation is multiplied by $\hat{v}$ (Eq. \ref{eq:Eq1}) to obtain the self- or cross-domain attention maps with a dimension of $[e, h\times w, h\times w]$. Finally, we average the outputs of the multiple heads for each token, pass the tensor through convolutional, batch normalization, and ReLU layers, and obtain $A_{attn} \in R^{h \times w \times c'}$.

\emph{\textbf{Cross-Domain Attention:}}
In order to emphasize the relationships between images originating from distinct domains, we introduce cross-domain attention with an unidirectional scheme from the source to the target domain, focusing on optimizing outcomes within the target domain. Unlike a bidirectional scheme, where the impact of both domains is equivalently considered through the computed cross-domain attention maps, our scheme lets the model to prioritize the characteristics of the target domain. To this end, first, the last-layer feature maps of the fourth scale from the source ($\mathcal{F}_{4,2}^s$) and target ($\mathcal{F}_{4,2}^t$) domains are added together to generate $\hat{v}$ (Fig. \ref{fig:Module_1}). In addition, we consider the source domain feature maps as $\hat{k}$ and the target domain feature maps as $\hat{q}$ for computing the cross-domain attention maps (Fig. \ref{fig:Module_1}). We perform DAB on feature pairs ($\mathcal{F}_{i,l}^s$ and $\mathcal{F}_{i,l}^t$) for each intermediate layer $l$ of a specific scale $i$.  
The output is then upsampled using bilinear interpolation and is used as the $\hat{v}$ of the next DAB, along with $\mathcal{F}_{i+1,l}^s$ and $\mathcal{F}_{i+1,l}^t$ being the new $\hat{q}$ and $\hat{k}$. {This process is repeated each time the DAB is used, for a total of three times, resulting in a collection of intermediate cross-domain attention maps.} Finally, after applying the last upsampling layer and achieving the $A_{attn}^{'}$, the DEB is applied to compute the tree density map (Fig.~\ref{fig:Module_1}). Within this DEB, the $\mathcal{F}_{1,2}^s$ and $\mathcal{F}_{1,2}^t$ as the last-layer feature maps of the first scale from the source and target domains are concatenated by a skip connection (Fig.~\ref{fig:Module_3}b). Subsequently, the combined result in DEB undergoes three consecutive convolutional and ReLU operations to reduce the number of channels to 1 and generate the tree density map. Overall, the cross-domain attention is used in the source-target subnet to estimate the $\mathcal{T}_{st}$ using $\mathcal{F}_{i,l}^s$ and $\mathcal{F}_{i,l}^t$.

\textbf{Self-Domain Attention:} The purpose of computing self-domain attention maps is to focus on relationships between different regions of an image within a domain. The structure of the self-domain attention is similar to the cross-domain attention except that it is computed within one domain by taking the $\hat{q}$ and $\hat{k}$ from the same domain at each scale ($\mathcal{F}_{i,l}^s$ or $\mathcal{F}_{i,l}^t$) instead of two different domains.  Utilizing the self-domain attention in the source and target subnets leads to generating the $\mathcal{T}_s$ and $\mathcal{T}_t$.

\subsubsection{Domain Discriminator}\label{sec:DomainDiscriminator}
In order to further enhance generalizability across different domains, our network is trained using an adversarial approach where an additional tree domain discriminator is involved. The role of the tree domain discriminator is to determine whether an image belongs to the target domain or source domain. Our domain discriminator architecture follows the structure of the classic VGG16 network (\cite{simonyan2014very}) {that takes the $\mathcal{T}_{s}$ and $\mathcal{T}_{t}$} as input. It consists of 16 layers, including 13 convolutional layers and 3 fully connected layers (Fig.~\ref{fig:overview}b). Each convolutional layer is followed by a ReLU activation function and uses $3\times3$ filters for convolution. The discriminator also incorporates max pooling layers to reduce the spatial resolution of feature maps. The flattened feature maps are then passed through fully connected layers. A softmax activation function is used in the last layer to assign probabilities to source and target classes. Further elaboration on the loss function is provided in Section \ref{sec:AdversarialLearning}.

\subsection{Domain Adaptive Learning strategy}\label{sec:LearningStrategy}

The learning strategy consists of three components: tree distribution matching (TDM), cross-domain feature alignment, and adversarial learning. The TDM involves optimizing network parameters by comparing the number of trees present in the estimated density map with the ground truth. The cross-domain feature alignment aligns the source and target domains by ensuring the similarity of the extracted self- and cross-domain attention {score maps}. Lastly, adversarial learning helps in making the feature distributions of the source and target domains similar.

\subsubsection{Tree Distribution Matching}\label{sec:TreeDistributionMatching}
The main idea of the TDM is to treat tree counting as a distribution matching problem (\cite{wang2020distribution}). Instead of directly estimating the tree density map, the TDM focuses on matching the distribution of the predicted tree density map with the ground truth distribution. This loss function incorporates a combination of the counting loss ($L_{count}$), optimal transport loss ($L_{ot}$), and total variation loss ($L_{tv}$). The $L_{count}$ measures the absolute difference between the estimated and ground truth tree numbers:
\begin{equation}
    \ L_{count} = \sum_{k=1}^K |\Vert \mathcal{S}_{k} \Vert - \Vert \mathcal{G}_k \Vert|  
\end{equation}
where $\mathcal{S} \in \{\mathcal{T}_{s}, \mathcal{T}_{t}, \mathcal{T}_{st} \}$ contains the estimated density maps and $\mathcal{G} \in \{ \mathcal{T}_{s}^{gt}, \mathcal{T}_{t}^{gt}, \mathcal{T}_{t}^{gt} \}$ contains their corresponding ground truth map, respectively;
$K=3$ and $\Vert. \Vert$ refers to the L1 norm used to aggregate the density values. 
The $L_{ot}$ is inspired by the optimal transport theory (\cite{arjovsky2017wasserstein}), which is a mathematical approach used to measure the dissimilarity between two different distributions. The $L_{ot}$ Loss compares the distribution of the predicted tree density map with the distribution of the ground truth by considering the optimal transportation plan between the two. It is calculated using the optimal transport cost as described in (\cite{wang2020distribution}):

\begin{equation}
    \ L_{ot} = \sum_{k=1}^K  \psi(\frac{\mathcal{S}_{k}}{\Vert \mathcal{S}_{k} \Vert},\frac{\mathcal{G}_{k}}{\Vert\mathcal{G}_{k} \Vert})
\end{equation}

where $\psi$ is the optimal transport cost (\cite{wang2020distribution}) that quantifies the dissimilarity between two probability distributions. This cost measures the minimal effort to transform the tree count distribution estimated by our model to the reference distribution, accounting for both the number of tree count adjustments and the distance required to relocate these trees. The $L_{tv}$ is a measure of the overall variation or complexity of an image by taking the sum of differences between the distribution of the estimated tree density map and the ground truth. This loss is applied to address the poor approximation of $L_{ot}$ in regions of low density (\cite{wang2020distribution}). 
\begin{equation}
    \ L_{tv}=\sum_{k=1}^K \frac{1}{2} \Vert \frac{\mathcal{S}_{k}}{\Vert \mathcal{S}_{k} \Vert}-\frac{\mathcal{G}_{k}}{\Vert \mathcal{G}_{k} \Vert}\Vert
\end{equation}

The overall tree distribution ($L_{TDM}$) matching loss for pixel-level learning is formulated as:

\begin{equation}
    \ L_{TDM}= \phi _1 L_{count}+\phi _2 L_{ot}+\phi _3 L_{tv}
\end{equation}
where $\phi _1$, $\phi _2$, and $\phi _3$ are the weight values that are set to 1, 0.1, and 0.01 according to (\cite{wang2020distribution}), respectively.

\subsubsection{Hierarchical Cross Domain Feature Alignment}\label{sec:HierarchicalCrossDomainFeatureAlignment}
The purpose of the hierarchical cross domain feature alignment (HCDFA) is to boost the robustness of our model's predictions when tested on previously unseen images by sharing knowledge across domains via feature alignment.
A sequential feature alignment at three scales of the decoder is employed to progressively align the source and target sequence features. As shown in Fig.~\ref{fig:overview}a, the proposed framework has three different subnets, in which the self-domain attention {score maps} ($A_{SDASM}$) are extracted from the first and last subnets while the cross-domain attention maps ($A_{CDASM}$) from the middle subnet (see Fig. \ref{fig:overview}c). The close distance between the self- and cross-domain attention {score maps} would enforce the model to generate domain-invariant feature maps, agnostic to any specific domain. This introduced loss function ($L_{HCDFA}$) is described as follows:

\begin{equation}
    \ L_{DS}= \sum\limits_{i=2}^{4}{[A_{CDASM}(i) - A_{SDASM}^{s} (i)]^2}
\end{equation}
\begin{equation}
    \ L_{DT}= \sum\limits_{i=2}^{4}{[A_{CDASM} (i) - A_{SDASM}^{t}(i)]^2}
\end{equation}

\begin{equation}
    \ L_{HCDFA}= \beta_1 L_{DS} + \beta_2 L_{DT} 
\end{equation}
where $L_{DS}$ represents the distance between the cross-domain attention {score maps} ($A_{CDASM}(i)$) and the source domain attention {score maps} ($A_{SDASM}^s(i)$), while $L_{DT}$ represents the distances between the cross-domain ($A_{CDASM}$(i)) and the target domain 
($A_{SDASM}^t(i)$)
attention score maps. The attention score maps are obtained following the aforementioned sections (Sec. \ref{sec:AttentiontoAdaptMechanism}); $i$ represents the scale of the corresponding encoder feature maps ($i \in \{2,3,4\}$) that is used.
$\beta_1$ and $\beta_2$ are the weight values for $L_{DS}$ and $L_{DT}$ respectively. Computing a weighted average of the $L_{DT}$ and $L_{DS}$ is critical to strike a balance between the source and target domains to obtain optimal results.

\subsubsection{Adversarial Learning}\label{sec:AdversarialLearning}
Referring to Sec.~\ref{sec:DomainDiscriminator}, our framework consists of two parts: the tree density generator ($\Phi_\text{G}$), which encompasses the shared encoder with the source and target subnets, and an additional tree domain discriminator network ($\Phi_\text{D}$) (Fig.~\ref{fig:overview}b). $\Phi_\text{G}$ generates tree density maps for input images, while $\Phi_\text{D}$ determines if an image is from the source or target domain. During the model's training, we train $\Phi_\text{G}$  to generate $\mathcal{T}_{s}$ from $\mathcal{I}_{s}$ using the source subnet and $\mathcal{T}_{t}$ from $\mathcal{I}_{t}$ using the target subnet, employing the supervised loss ($L_{TDM}$). Meanwhile, we train $\Phi_\text{D}$ {to assign high scores (1) to the $\mathcal{T}_{s}$ and low scores (0) to the $\mathcal{T}_{t}$.}
$\Phi_\text{D}$ learns to differentiate between the source and target domain data, while the $\Phi_\text{G}$ simultaneously aims to produce domain-invariant representations, thereby enhancing the model's adaptation to the target domain.
Given $m$ source domain training images $\mathcal{I}_t$ and $n$ target domain training images $\mathcal{I}_t$ with their corresponding ground truth ($\mathcal{T}_{s}$ and $\mathcal{T}_{t}$), we define the loss function following (\cite{zhang2017deep}):

\begin{equation}
    \ \min_{\theta_{\Phi_\text{G}}} \max_{\theta_{\Phi_\text{G}}} L_{Adv}(\theta_{\Phi_\text{G}},\theta_{\Phi_\text{D}})= \sum \limits _{i=1}^{m}{L_{TDM}^s}+\sum \limits _{i=1}^{n}{L_{TDM}^t}-\lambda[\sum \limits _{i=1}^{m}{L_{bce}(\Phi_\text{D}(\Phi_\text{G}(\mathcal{I}_s),\mathcal{I}_s),1)}+\sum \limits _{i=1}^{n}{L_{bce}(\Phi_\text{D}(\Phi_\text{G}(\mathcal{I}_t),\mathcal{I}_t),0)}]
\end{equation}

where, $\theta_{\Phi_\text{G}}$ and $\theta_{\Phi_\text{D}}$ are the parameters of the $\Phi_\text{G}$ and $\Phi_\text{D}$, respectively. $L_{bce}$ represents the binary-class cross-entropy loss, $L_{TDM}^s$ and $L_{TDM}^t$ represent the computed $L_{TDM}$ for source and target domain images. The terms of $\sum \limits _{i=1}^{m}{L_{TDM}^s}+\sum \limits _{i=1}^{n}{L_{TDM}^t}$ in the $L_{Adv}$ represent the supervised training of $\Phi_\text{G}$ using source and target data, while the remained terms represent the adversarial training part. During the training process, we minimize a part of the loss with respect to the parameters $\theta_{\Phi_\text{G}}$ of $\Phi_\text{G}$, while maximizing the loss with respect to the parameters $\theta_{\Phi_\text{D}}$ of $\Phi_\text{D}$.

\subsubsection{Training loss}\label{sec:TrainingLoss}
Overall, the final loss function is composed of three different losses. The first loss is based on the summation of the computed supervised loss ($L_{TDM}$) of the three subnets of the network including the source ($L_{TDM}^s$), target ($L_{TDM}^s$), and source-target ($L_{TDM}^{st}$) losses ($L_{TDM} = L_{TDM}^s + L_{TDM}^t + L_{TDM}^{st}$). The second and the third losses are the $L_{HCDFA}$ and $L_{Adv}$. The final loss is:
\begin{equation}
    \ L = L_{TDM} + L_{HCDFA} + L_{Adv}
\end{equation}

\subsubsection{Inference}\label{sec:Inference}
During the inference process, only the target subnet is employed, as depicted in Fig.~\ref{fig:overview}a.

\section{Experiments}\label{sec:Experiments}
\subsection{Datasets}\label{sec:Datasets}
We employ three different datasets (Fig.~\ref{fig:dataset}) which were gathered from different domains, \ie{different countries that contain various tree shapes, sizes, and distributions (Table.~\ref{tab:DatasetCharacteristics}).} The tree density map uses a jet colormap, where blue indicates low density and red indicates high density values.

The London, Jiangsu and Yosemite datasets have unique characteristics which are summarized in Table 1. They contain different tree types with different shapes, canopies, and densities that are in various landscapes. For example, the London and Jiangsu datasets contain residential areas, while the Yosemite dataset is gathered from wooded mountainous areas (Fig. 1). Moreover, the average tree number per image of the London, Jiangsu, and Yosemite datasets is 155, 276, and 36, respectively. 

\subsubsection{Jiangsu Dataset}\label{sec:JiangsuDataset}
This dataset encompasses 24 satellite images taken by the GaofenII satellite with a ground sample distance (GSD) of 0.8m (available at https://github.com/sddpltwanqiu/TreeCountNet/tree/main). These images were captured in Jiangsu Province, China (\cite{yao2021tree,liu2021deep}), which cover diverse landscapes such as urban residential, cropland, and hills. There are 664,487 manually annotated trees spread across 2400 images, 1920 images are used for training while 480 images for testing.

\subsubsection{London Dataset}\label{sec:LondonDataset}
This dataset consists of high-resolution images captured at 0.2m GSD from London, United Kingdom for training and testing (available at https://github.com/HAAClassic/TreeFormer/tree/main). Among the labeled images, 452 samples are designated for training and 161 samples are for testing. There are 95,067 trees that are manually annotated across 613 images. The images cover different landscapes such as urban residential areas and dense parks. This dataset is divided into a training set of 450 images and a test set of 152 images.

\subsubsection{Yosemite Dataset}\label{sec:YosemiteDataset}
The study area for this dataset revolves around Yosemite National Park, located in California, United States of America (available at \emph{https://github.com/nightonion/yosemite-tree-dataset}). It comprises a rectangular image from Google Maps with dimensions of 19,200 × 38,400 pixels and a GSD of 0.12m. Within this image, 98,949 trees have been manually annotated. {
The image is split into 2700 smaller images, with 1350 chosen for training and the remaining 1350 selected for testing.}

\begin{table*}[!t]
\centering
	\caption{Characteristics of the utilized datasets}
	\begin{center}
	\begin{tabular}{c|c c c c c}
   \toprule
		Dataset name &  Location & Landscape type & \thead{Number of \\ trees/image}  &  \thead{Total number \\of trees } \\[5pt]  
	\midrule
	    London & UK, Europe & Urban, residential area, dense park & 155 & 95,067 \\[3pt]
	    Jiangsu &  China, Asia & Cropland, suburban, urban, residential area  & 276 & 664,487\\[3pt]
	    Yosemite &  US, North America & Wooded mountainous & 36 & 98,949 \\[3pt]
	\bottomrule
	\end{tabular}
	\end{center}
    \label{tab:DatasetCharacteristics}	
\end{table*}

\begin{figure*}[!t]
\centering
\includegraphics[width=6.5 in]{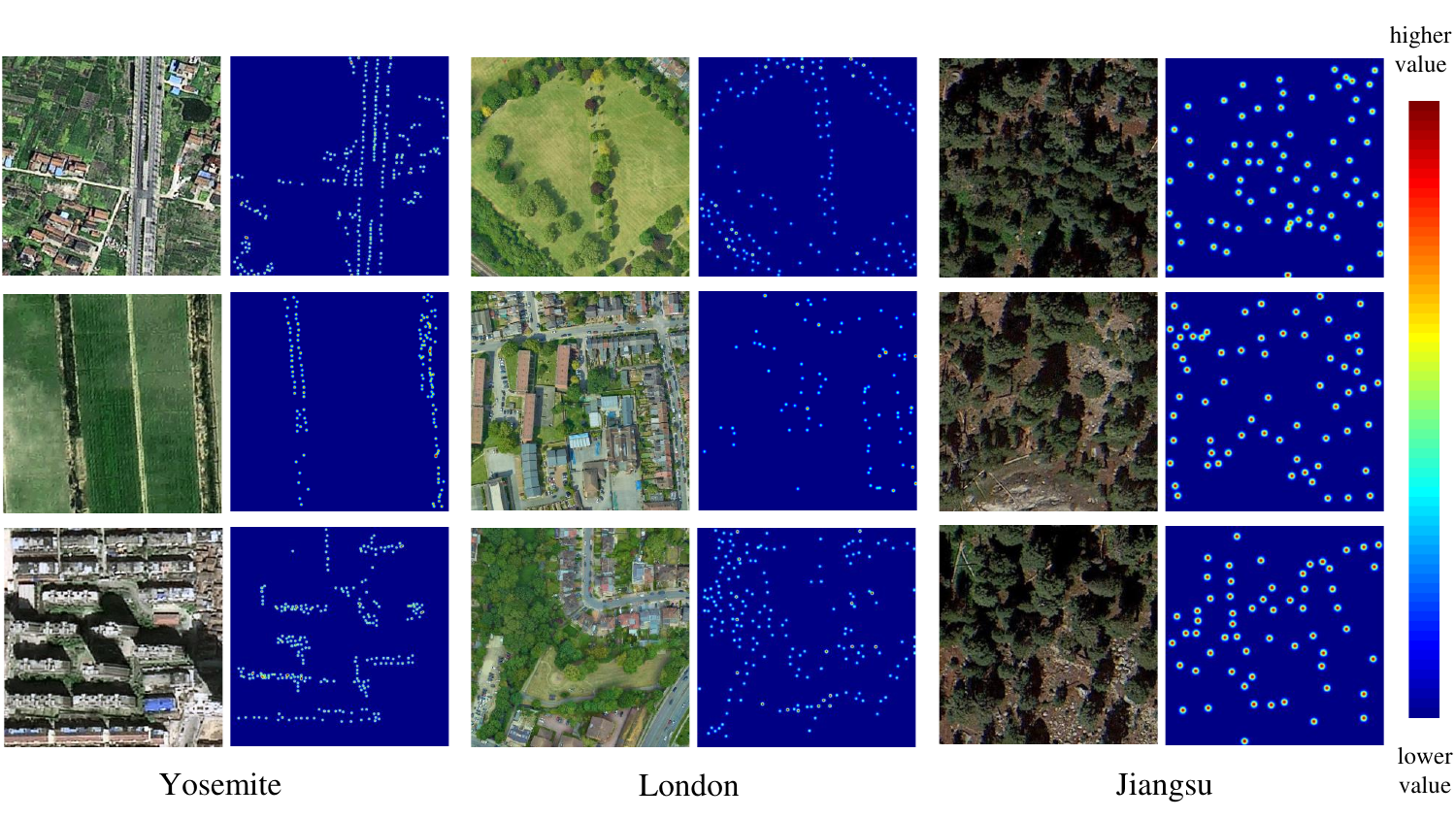}
\caption{Some samples of RGB images and corresponding tree density maps of the Yosemite, London, and Jiangsu datasets.}

\label{fig:dataset}
\end{figure*}

\subsection{Evaluation Metrics}\label{sec:EvaluationMetrics}
The estimated tree density maps are evaluated using two types of criteria. In the first type, the number of trees in the estimated density map is compared with the reference data using three metrics including mean absolute error ($E_{MAE}$), root mean squared error ($E_{RMS}$), and R-Squared ($E_{R2}$) (\cite{yao2021tree,chen2022transformer}). They are defined as follows:
\begin{equation}
    \ E_{MAE}=\frac{1}{N}\sum \limits _{i=0}^N |y_i^\text{es}-y_i^\text{gt}|
\end{equation}

\begin{equation}
    \ E_{RMS}=\sqrt{\frac{1}{N}\sum \limits _{i=0}^N (y_i^\text{es}-y_i^\text{gt})^2}
\end{equation}

\begin{equation}
    \ E_{R^2}=1-\frac{\sum \limits _{i=0}^N (y_i^\text{es}-y_i^\text{gt})^2}{\sum \limits _{i=0}^N (y_i^\text{es}-\bar{y}^\text{gt})^2}
\end{equation} 
where $y_i^\text{es}$ represents the estimated tree number for the $i$-th image, $y^i_{gt}$ is the corresponding ground truth tree number. The $\bar{y}^\text{gt}$ is the mean ground truth tree number over images and $N$ denotes the number of images. In general, higher $E_{R2}$ and lower $E_{RMS}$ and $E_{MAE}$ values indicate better performance. 

In the second type, the location of detected trees is assessed using reference data by three metrics including the Precision ($E_P$), Recall ($E_R$), and F1-measure ($E_{F_1}$)(\cite{wang2020nwpu}). They are calculated based on the number of true positives, false positives, and false negatives obtained from the comparison between the predicted density map and the ground truth density map. Following (\cite{wang2020nwpu, liu2019point}), for each point, a search circle with a radius of 15 pixels is considered {as the assumed covered area by a tree crown,} and the following parameters are computed based on the search circle:
\begin{itemize}
\item True Positives ($TP$): This refers to the number of trees that have been correctly detected. A correctly detected tree is defined as the closest detected tree within the search circle of each ground truth (GT) point.
\item False Positives ($FP$): This represents the number of trees that have been incorrectly detected. Any detected tree that does not fall within the search circle of any ground truth point is considered an incorrect detection.
\item False Negatives ($FN$): This indicates the number of trees that have been missed in detection. When no trees are found within the specified search circle of a given ground truth point, it is considered a missed detection. 
\end{itemize}
The $E_P$, $E_R$, and $E_{F1}$ are defined following (\cite{wang2020nwpu}):
\begin{equation}
    \ E_p=TP/(FP+TP)
\end{equation}
\begin{equation}
    \ E_R=TP/(FN+TP)
\end{equation}
\begin{equation}
    \ E_{F1}=2(E_p \times E_R )/(E_p+E_R )
\end{equation}
In general, higher $E_{P}$, $E_{R}$ and $E_{F1}$ values indicate better performance.

\subsection{Implementation Details}\label{sec:ImplementationDetails}
The parameters of the transformer are set according to (\cite{liu2021swin}).
To enhance the training set, we apply horizontal flipping and random cropping to source and target domain images (\cite{amirkolaee2019convolutional,amirkolaee2022development}). Additionally, the input images are randomly cropped to a fixed size of $256 \times 256$ for network input. We only select 5 labeled images (5-shot) from the target domain to evaluate the cross-domain capability of AdaTreeFormer by default. 5-shot learning is a common benchmark practice so as to strike a balance between using few shots and providing essential information for the model The number of batch size, epochs, weight decay, and learning rate are set to 8, 200, $10^{-5}$ and $10^{-4}$, respectively. We use the Adam optimizer. To optimize the adversarial loss function ($L_{Adv}$), we use a standard stochastic gradient descent method. Initially, we set $\lambda$ to 0.1 and increase it to 1 after 100 epochs when {tree density generator $\Phi_{\text G}$} starts producing satisfactory tree density maps. {As the London dataset images possess high spatial resolution and encompass diverse areas with varying types and densities of tree cover,} all parameters are tuned on the London dataset and applied to all experiments. The ground truth consists of the coordinates of tree locations, indicated by annotation dots. We generate the ground truth density maps from the tree locations using Gaussian functions following (\cite{zhang2016single}).

\subsection{Comparisons with state of the art}\label{sec:ComparisonsWithStateOfThe}

The performance of the proposed AdaTreeFormer is investigated in six different scenarios, considering the study areas of London ($S_L$), Jiangsu ($S_J$), and Yosemite ($S_Y$). In each scenario, one dataset is set as the source domain, and the other as the target domain. For example, in $S_L \rightarrow S_J$, all images of the $S_L$ along with 5 shots of images of $S_J$ are considered for training the network.

In order to compare to other existing methods, the results of different networks are presented in Table \ref{tab:CompareSOTA}.  AdaCrowd (\cite{reddy2021adacrowd}), CODA (\cite{li2019coda}), CycleGAN (\cite{zhu2017unpaired}), SE-CycleGAN (\cite{wang2004image}), and DGCC (\cite{du2023domain}) utilize unsupervised domain adaptation techniques where {the entire unlabeled images (the whole training set) of the target domain are used for adaptation}, while the NLT (\cite{wang2021neuron}) and FSCC (\cite{reddy2020few}) employ few shots (5 shots by default) of the target domain {with their labels} for domain adaptation. Moreover, AdaTreeFormer can be degraded with one subnet for training on the source data only, which is indicated as the Baseline in the table; this trained baseline can be further fine-tuned (FT) using 5 shots of the target domain (Baseline+FT). In addition, the performance of cross-head supervision network (CHSNet) (\cite{dai2023cross}) and FusionCount (\cite{yiming2022fusioncount}) as conventional supervised methods are also fine-tuned using 5 shots in the target domain.

According to Table \ref{tab:CompareSOTA}, the NLT approach (\cite{wang2021neuron}) in general achieves the second position in the designed scenarios for few-shot domain adaptation. In $S_J \rightarrow S_L$ scenario, the NTL has only 2.3 and 2.9 decreases of $E_{MAE}$ and $E_{RMS}$ compared to the FSCC as the previously best-performed model, while our model achieves significantly larger reductions of 6.7 and 5.3 in $E_{MAE}$ and $E_{RMS}$, respectively. The achieved $E_{MAE}$ using our model in the $S_Y \rightarrow S_J$ scenario is 15.9, much lower than that of the NLT, demonstrating a significant improvement in tree number estimation. The $E_{RMS}$ puts more weight on larger errors compared to the $E_{MAE}$. The $E_{RMS}$ of NLT in the mentioned scenario is 68.7, much larger than that of our model, which indicates a drastic reduction of the blunders in our model. The  $E_{R2}$ represents the proportion of variance in the actual tree counts. It has a range from 0 to 1, and a value of 1 means the model perfectly explains the variation in actual tree counts. The NLT obtains 0.5 of $E_{R^2}$ while our model achieves 0.6. 
The $E_P$, $E_R$, and $E_{F1}$ are used to assess the location of the detected trees in the estimated density map. The $E_P$ and $E_R$ focus on correctness and completeness of the tree locations in the estimated tree density maps, while the $E_{F1}$ combines these for a balanced view. For example, in $S_J \rightarrow S_L$ scenario, an increase of 3.1, 5.2 and 4.2 is observed for $E_P$, $E_R$, and $E_{F1}$ from FSCC to NLT, while our model achieves significantly larger gains of 12.8, 18.9 and 15.9 for these metrics. In $S_Y \rightarrow S_J$ scenario, the $E_P$, $E_R$, and $E_{F1}$ obtained by our model are 19.3\%, 19.1\%, and 23.7\% more than the NLT, signifying a substantial improvement by our model.

The differences in the datasets (Fig~\ref{fig:dataset}) cause differences in challenges faced by each of the designed scenarios. For example, in the $S_Y \rightarrow S_J$ scenario, since the Yosemite dataset ($S_Y$) does not cover croplands or cities, it is challenging to extract this knowledge from only a few shots of the labeled images in Jiangsu dataset ($S_J$).
Generally, the more diverse in tree type, shape, size, canopy, and density the source domain is, the better the network is trained and produces better results in the target domain. Whereas, in the case of training the model with less variety of data characteristics in the source domain, it becomes very difficult to achieve high accuracy in the target domain. Despite the effective performance of our proposed framework in domain adaptation, the impact of changing the source and target domains in the designed scenarios can still be seen. For example, the achieved $E_{MAE}$ in the $S_Y \rightarrow S_L$ scenario is 1.7 more than that in the $S_J \rightarrow S_L$ scenario, owing to the more diversity of trees and landscapes in the Jiangsu dataset ($S_J$) compared to the Yosemite dataset ($S_Y$).

\begin{table*}[!t]
\centering
	\caption{Comparison with the state of the art on the London, Jiangsu, and Yosemite datasets. The best and second results are marked in red and blue, respectively.}
	\begin{center}
    \scalebox{0.75}{
	\begin{tabular}{c c c|c c c c c c|c c c c c c}
   \toprule
	\midrule
		{} & {} & {} & $E_{MAE} \downarrow$ & $E_{RMS}  \downarrow$ & $E_{R^2} \uparrow$ & $E_{P} \uparrow $ & $E_{R} \uparrow$  & $E_{F1} \uparrow$& $E_{MAE} \downarrow$ & $E_{RMS}  \downarrow$ & $E_{R^2} \uparrow$ & $E_{P} \uparrow $ & $E_{R} \uparrow$  & $E_{F1} \uparrow$\\
    \midrule
	   {Method} & Adpt & FS & \multicolumn{6}{c}{$S_L \rightarrow S_J$} & \multicolumn{6}{c}{$S_L \rightarrow S_Y$}\\
     
	\midrule 
        AdaCrowd & \cmark & \xmark & 243.6 & 399.6 & -0.5 & 5.6 & 12.0 & 7.6 & 26.1 & 49.9 & -0.6 & 20.4 & 15.3 & 17.5 \\
	  CODA & \cmark & \xmark & 239.3 & 359.6 & -0.3 & 41.9 & 1.6 & 3.2 & 24.7 & 32.5 & -0.1 & 32.5 & 23.9 & 27.6\\
	  CycleGAN & \cmark & \xmark & 284.2 & 468.5 & -1.1 & 32.8 & 2.6 & 4.9 & 28.9 & 36.3 & -0.6 & 22.3 & 24.1 & 23.2 \\  
   	SE-CycleGAN & \cmark & \xmark & 268.3 & 432.5 & -1.0 & 24.6 & 4.1 & 7.1 & 21.8 & 26.5 & 0.1 & 38.4 & 23.6 & 29.2\\
        DGCC & \cmark & \xmark &  192.4 & 350.4 & -0.1 & 24.5 & 18.3 & 21.0 & 17.4 & 21.2 & 0.03 & 24.2 & 19.4 & 21.5 \\
        NLT & \cmark & \cmark &  137.0 & 194.5 & 0.4 & 58.1 & \textcolor{blue}{39.5} & \textcolor{blue}{47.0} & 12.1 &  \textcolor{blue}{14.6} & \textcolor{blue}{0.6} & 55.3 & \textcolor{blue}{63.7} & \textcolor{blue}{59.2}\\
	  FSCC & \cmark & \cmark & 176.1 & 356.2 & -0.2 & 18.3 & 25.4 & 21.3 & 16.3 & 19.4 & 0.4 & 39.5 & 31.4 & 35.0 \\
	  FusionCount+FT  & \xmark & \cmark & 145.8 & 203.2 & 0.4 & 65.1 & 14.3 & 23.5 & 14.3 & 18.7 & 0.4 & 11.2 & 47.7 & 18.1\\ 
        CHSNet+FT  & \xmark & \cmark & 266.1 & 414.1 & -0.6 & \textcolor{red}{96.9} & 0.2 & 0.4 & 22.1 & 25.6 &  0.1 & \textcolor{blue}{58.3} & 8.0 & 14.1\\ 
        Baseline & \xmark & \xmark & 273.0 & 421.6 & -0.6 & 43.9 & 2.7 & 5.2 &  25.7 & 29.8 & 0.01 & 35.7 & 24.1 & 28.8 \\
	  Baseline+FT & \xmark & \cmark & \textcolor{blue}{134.0} & \textcolor{blue}{181.5} & \textcolor{blue}{0.5} & 66.7 & 33.4 &  44.5 & \textcolor{blue}{11.5} & 15.7 & 0.6 & 52.3 & 61.8 & 56.7\\
	  AdaTreeFormer & \cmark & \cmark & \textcolor{red}{122.5} & \textcolor{red}{168.4} &  \textcolor{red}{0.7} & \textcolor{blue}{75.4} & \textcolor{red}{41.8} &  \textcolor{red}{53.8} & \textcolor{red}{7.7} & \textcolor{red}{10.8} &  \textcolor{red}{0.7} & \textcolor{red}{64.3} & \textcolor{red}{68.7} &  \textcolor{red}{66.4}\\ 
    \midrule
	   {} & & & \multicolumn{6}{c}{$S_J \rightarrow S_L$} & \multicolumn{6}{c}{$S_J \rightarrow S_Y$}\\
	\midrule 
        AdaCrowd & \cmark & \xmark & 110.2 & 143.9 & -4.4 & 33.1 & 46.3 & 38.6 & 26.8 & 34.5 & -0.1 & 22.7 & 25.3 &  24.0 \\
	  CODA & \cmark & \xmark & 86.6 & 103.9 & -1.4 & 38.5 & 14.6 & 21.2 & 22.7 & 29.4 & 0.1 & 35.4 & 27.1 & 30.7 \\
	  CycleGAN & \cmark & \xmark & 123.6 & 146.3 & -7.5 & 32.5 & 42.7 & 36.9 & 29.2 & 36.1 & -0.5 & 29.5 & 30.6 & 30.1 \\  
   	SE-CycleGAN & \cmark & \xmark & 105.8 & 126.4 & -2.5 & 38.6 & 43.2 & 40.8 & 23.5 & 28.7 & -0.1 & 35.2 & 32.7 & 33.9 \\  
        DGCC & \cmark & \xmark & 50.7  & 63.0 & 0.04 & 32.5 & 21.4 & 25.8 & 19.4 & 24.0 & 0.1 & 10.3 & 23.9 & 14.4 \\  
        NLT & \cmark & \cmark & \textcolor{blue}{25.8} & \textcolor{blue}{30.7} & \textcolor{blue}{0.5} & \textcolor{blue}{61.4} & \textcolor{blue}{58.3} & \textcolor{blue}{59.8} & \textcolor{blue}{11.4} & \textcolor{blue}{14.9} & \textcolor{blue}{0.6} & 59.6 & 58.2 & 58.7\\
	  FSCC & \cmark & \cmark & 28.1 & 33.6 & 0.5 & 58.3 & 53.1 & 55.6 & 12.2 & 18.3 & 0.4 & \textcolor{blue}{59.6} & \textcolor{blue}{67.4} & \textcolor{blue}{63.2} \\
        FusionCount+FT & \xmark & \cmark & 36.3 & 59.1 & 0.01 & 55.2 & 2.4 & 4.6 & 13.2 & 16.4 & 0.5 & 32.3 & 45.2 & 37.7\\ 
        CHSNet+FT & \xmark & \cmark & 30.1 & 40.2 & 0.2 & 50.9 & 14.8 & 23.0 & 12.9 & 15.6 & 0.6 & 48.0 & 23.4 & 31.5\\ 
	  Baseline & \xmark & \xmark & 124.7 & 165.5 & 0.1 & 30.7 & 45.3 & 36.6 & 27.4 & 32.5 & -0.2 & 31.4 & 22.6 & 25.5\\
	  Baseline+FT & \xmark & \cmark & 29.4 & 36.3 & 0.4 & 56.7 & 59.6 & 58.1 & 13.5 & 16.4 & 0.5 & 56.2 & 49.6 & 52.7 \\ 
	  AdaTreeFormer & \cmark & \cmark & \textcolor{red}{21.4} & \textcolor{red}{28.3} &  \textcolor{red}{0.6} & \textcolor{red}{71.1} & \textcolor{red}{72.0} &  \textcolor{red}{71.5}& \textcolor{red}{8.9} & \textcolor{red}{11.1} &  \textcolor{red}{0.7} & \textcolor{red}{61.9} & \textcolor{red}{69.5} &  \textcolor{red}{65.5}\\
    \midrule
	   {} & & & \multicolumn{6}{c}{$S_Y \rightarrow S_L$} & \multicolumn{6}{c}{$S_Y \rightarrow S_J$}\\
	\midrule 
        AdaCrowd & \cmark & \xmark & 129.4 & 142.4 & -4.2 & 35.6 & 10.4 & 16.2 & 182.7 & 224.5 & 0.2 & 43.2 & 24.5 & 31.3 \\
	  CODA & \cmark & \xmark & 105.3 & 124.8 & -2.3 & 33.2 & 12.4 & 18.1 & 214.8 & 334.4 & 0.02 & 36.4 & 19.4 & 25.1  \\
	  CycleGAN & \cmark & \xmark & 134.5 & 152.7 & -8.3 & 30.1 & 9.72 & 14.9 & 242.8 & 372.4 & -0.01 & 29.2 & 15.3 & 20.1 \\ 
   	SE-CycleGAN & \cmark & \xmark & 112.8 & 125.6 & -3.5 & 34.6 & 12.8 & 18.7 & 224.9 & 410.5 & -0.1 & 28.4 & 18.3 & 22.3\\  
        DGCC & \cmark & \xmark  & 62.5 & 76.1 & -0.6 & 43.6 & 37.2 & 20.0 & 214.6 & 372.4 & 0.04 & 12.5 & 19.3  & 15.2\\ 
        NLT & \cmark & \cmark & \textcolor{blue}{34.2} & 41.6 & 0.2 & 60.4 & \textcolor{blue}{64.1} & \textcolor{blue}{62.2} & 143.5 & 205.2 & 0.49 & 62.1 & 25.4 & 36.1\\
	  FSCC & \cmark & \cmark & 36.5 & 42.2 & 0.2 & 61.2 & 59.2 & 60.3 & 163.2 & 197.6 & 0.5 & 63.7 & 22.4 & 33.2 \\
	  FusionCount+FT  & \xmark & \cmark & 56.3 & 68.4 & 0.02 & 42.4 & 10.6 & 16.9 & \textcolor{blue}{138.7} & 239.0 & 0.5 & 58.2 & 14.5 & 23.2\\ 
        CHSNet+FT  & \xmark & \cmark &33.2 & 56.7 & 1.3 & \textcolor{blue}{69.0} & 4.1 & 7.7 &  216.4 & 382.9 & -0.3 &  22.9 & 0.5 & 1.0\\ 
        Baseline & \xmark & \xmark & 122.6 & 129.8 & -6.8 & 39.0 & 9.5 & 15.3 & 271.4 & 422.1 & -0.7 & 50.0 & 2.4 & 4.7\\
	  Baseline+FT & \xmark & \cmark & 34.5 & \textcolor{blue}{39.3} & \textcolor{blue}{0.2} & 58.4 & 60.2 & 59.3 & 149.4 & \textcolor{blue}{197.3} & \textcolor{blue}{0.5} & \textcolor{blue}{69.2} & \textcolor{blue}{27.4} & \textcolor{blue}{39.3} \\
	  AdaTreeFormer & \cmark & \cmark & \textcolor{red}{23.1} & \textcolor{red}{31.0} &  \textcolor{red}{0.5} & \textcolor{red}{69.3} & \textcolor{red}{72.4} &  \textcolor{red}{70.8} & \textcolor{red}{122.8} & \textcolor{red}{170.3} &  \textcolor{red}{0.6} & \textcolor{red}{77.5} & \textcolor{red}{33.6} &  \textcolor{red}{46.9}\\ 
    \midrule
	\bottomrule
	\end{tabular}
 }
	\end{center}
    \label{tab:CompareSOTA}	
\end{table*}

\begin{figure}[htbp!]
\centering
\includegraphics[width=5.5in]{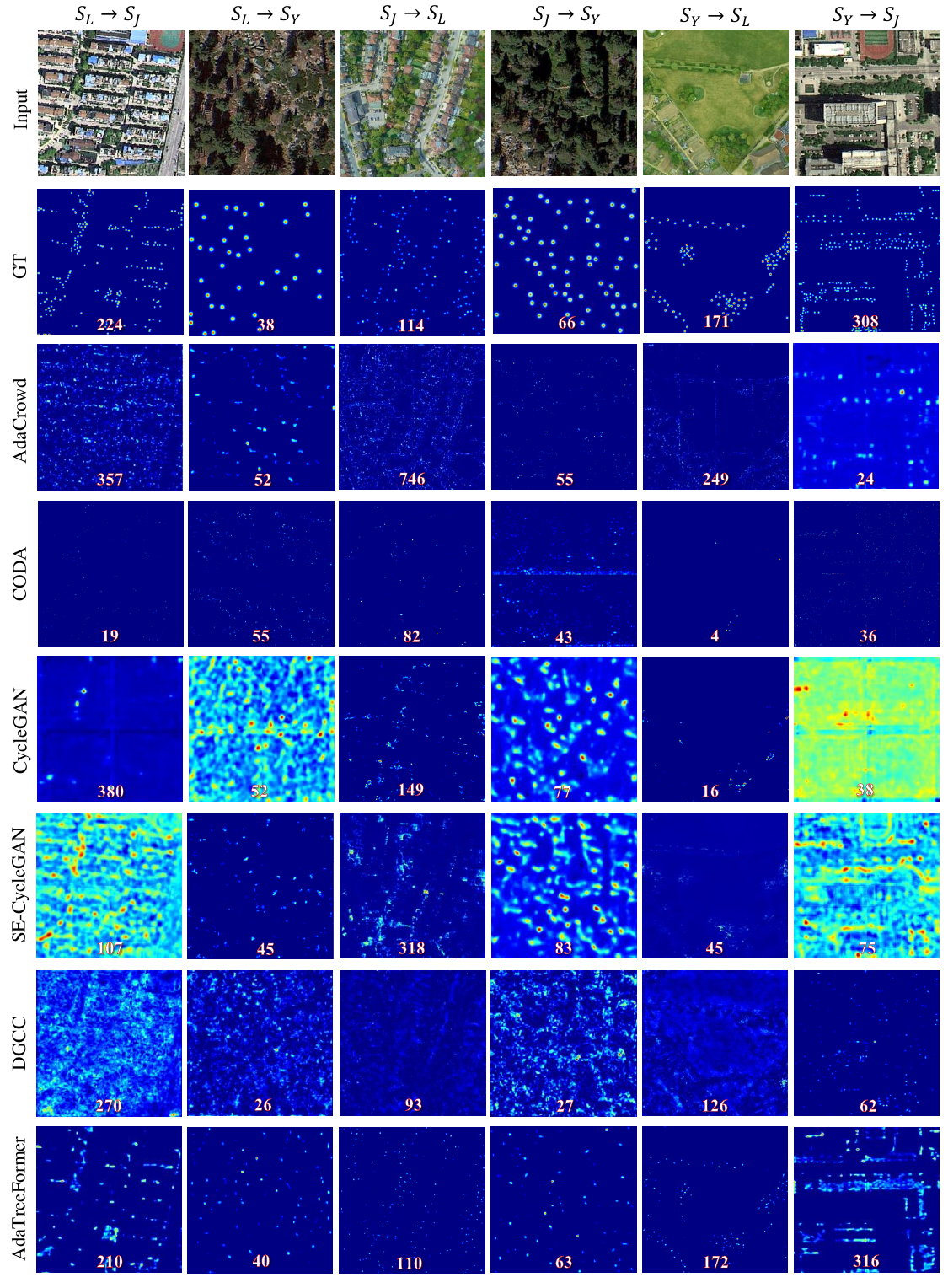}
\caption{Qualitative results of AdaTreeFormer compared with the state of the art methods that exclusively rely on adaptation techniques including Adacrowd, CODA, CycleGAN, SE-CycleGAN, and DGCC on London, Jiangsu, and Yosemite datasets. The first row shows sample images of three utilized datasets. The remaining rows show density maps of GT and other methods.}

\label{fig:SOTA1_results}
\end{figure}

\begin{figure}[htbp!]
\centering
\includegraphics[width=5.5in]{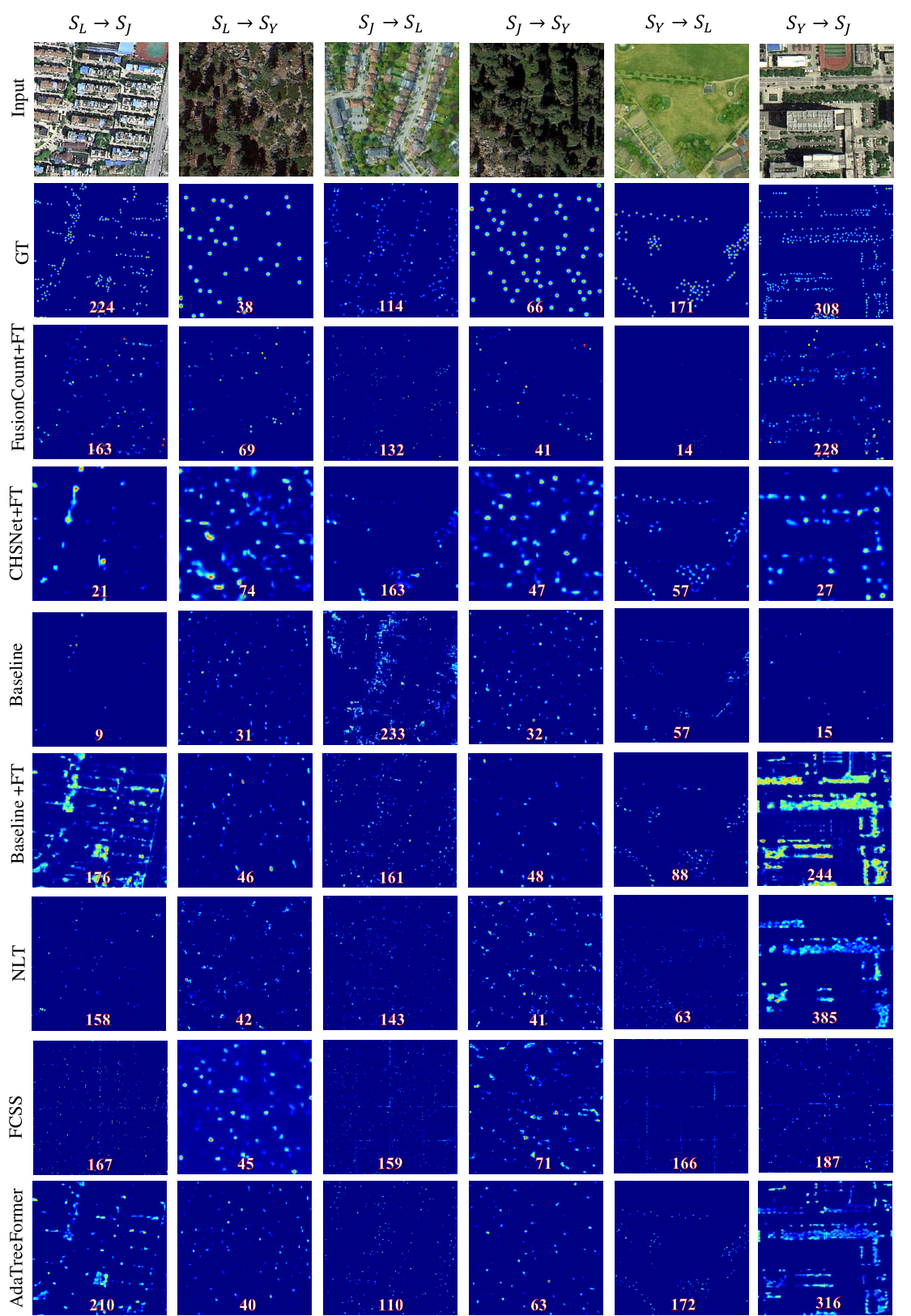}
\caption{Qualitative results of AdaTreeFormer compared with the state of the art methods that incorporate only a few shots from the target domain during the training process including FusionCount+FT, CHSNet+FT, Baseline, Baseline+FT, NLT, and FCSS on London, Jiangsu, and Yosemite datasets. The first row shows sample images of three utilized datasets. The remaining rows show density maps of GT and other methods.}

\label{fig:SOTA2_results}
\end{figure}

According to the Table \ref{tab:CompareSOTA}, methods that fall within {unsupervised} domain adaptation do not perform well in the target domains (AdaCrowd, CODA, CycleGAN, SE-CycleGAN, and GDCC). Furthermore,  methods that employ a few images related to the target domain for simple fine-tuning also fail to achieve the desired accuracy in their results (FusionCount+FT and CHSNet+FT). Whereas methods that use both adaptation techniques and a few shots of the target domain simultaneously perform better and achieve more accurate results (NLT and FSCC). Moreover, since the domain shift is generally high and only using image translation algorithms (CycleGAN and SE-CycleGAN), subdomain division (GSCC), parameter tuning (AdaCrowd), or self-supervised loss (CODA) do not achieve good performance in the target domain.

In Figure \ref{fig:SOTA1_results}, we present qualitative results of our method compared to other methods that exclusively rely on adaptation techniques. It is evident that our approach outperforms these methods in terms of quality and accuracy. Additionally, in Figure \ref{fig:SOTA2_results}, we showcase the results of methods that incorporate only a few shots from the target domain during the training process. 
Our method demonstrates superior performance in this regard. 

Of course, no method is guaranteed to work in every scenario, and always has its limitations. Some factors can also hinder the performance of our network. For example, when the source and target domain images pertain to distinct seasons, the performance of the model might deteriorate. Indeed, if the target domain data has leafless trees in winter while the source domain data has leafy trees in summer, good results will not be obtained in the target domain. Also, if the density of trees in the source domain is very sparse and the diversity of the landscape is very low, but the target area has a complex structure, the performance of the model will be reduced. However, we believe that our model already significantly advances domain adaptation for tree counting by outperforming existing methods, providing insights for future works along this research line.

\subsection{Ablation Study}\label{sec:AblationStudy}
We analyze AdaTreeFormer in the $S_L \rightarrow S_J$, $S_J \rightarrow S_L$, and $S_Y \rightarrow S_L$ domain adaptation scenarios by ablating its proposed components to evaluate their effects on the model accuracy. We employ the $E_{MAE}$ and $E_{F1}$ for evaluation

\subsubsection{Analysis on model architecture}
In this section, we investigate the proposed multi-scale feature extraction and attention-to-adapt mechanism.

\textbf{Efficiency of Data Augmentation.} To analyze the performance of the AdaTreeFormer using the Cutmix as data augmentation for the target domain, the accuracy of the achieved results without using the Cutmix is computed (w/o Cutmix) which increases $E_{MAE}$ by 2.8, 2.5, and 2.2 and decreases $E_{F1}$ by 2.6\%, 2.7\% and 4.4\% for the mentioned scenarios in Table \ref{tab:HTFE}.  
Moreover, we can apply Mixup (w/ Mixup) (\cite{zhang2017mixup}) or multi-size cropping (\cite{van2021revisiting}) instead of Cutmix to increase the number of training images. In multi-size cropping, the input image undergoes random cropping with different resolutions, and the outcomes are subsequently resized to $256 \times 256$. Accordingly, replacing Cutmix with other ways has reduced the accuracy of the results.

\textbf{Hierarchical Tree Feature Extraction:}
The hierarchical structure of the HTFE can be downgraded by reducing the number of scales of the encoder from 4 into 2 (Scales 1 and 2 in Fig.~\ref{fig:Module_1}) so that only one scale is produced for the decoder (single scale). This scale reduction increases $E_{MAE}$ by 25.7, 6.9, and 7.4 and decreases $E_{F1}$ by 4.6\%, 13.1\% and 15.2\% for $S_L \rightarrow S_J$, $S_J \rightarrow S_L$, and $S_Y \rightarrow S_L$, respectively (Table \ref{tab:HTFE}). 

\begin{table}[htbp!]
\centering
	\caption{Ablation study of the employed feature extraction and data augmentation on the $S_L \rightarrow S_J$, $S_J \rightarrow S_L$, and $S_Y \rightarrow S_L$ domain adaptation scenarios.}
	\begin{center}
	\begin{tabular}{c|c c |c c | c c}
    \toprule
	\midrule
		Scenario & \multicolumn{2}{c}{$S_L \rightarrow S_J$} & \multicolumn{2}{c}{$S_J \rightarrow S_L$} &  \multicolumn{2}{c}{$S_Y \rightarrow S_L$} \\
    \midrule
		Method  & $E_{MAE}$ & $E_{F1}$ & $E_{MAE}$ & $E_{F1}$ & $E_{MAE}$ & $E_{F1}$ \\
    \midrule
	  {Single scale} & 148.2 & 49.2 & 28.3 & 58.4 & 30.5 & 55.6 \\  
        {w/o Cutmix} & 125.3 & 51.2 & 23.9 & 68.8 & 25.3 & 66.4 \\  
        {w/ Crop} & 127.4 & 50.8 & 24.3 & 67.2 & 27.8 & 63.7\\  
        {w/ Mixup} & 152.3 & 24.9 & 33.4 & 26.3 & 36.9 & 44.8\\  
        {AdaTreeFormer} & \textbf{122.5} & \textbf{53.8} & \textbf{21.4} & \textbf{71.5} & \textbf{23.1} & \textbf{70.8}   \\   
	\bottomrule
	\end{tabular}
	\end{center}
    \label{tab:HTFE}	
\end{table}

\textbf{Attention-to-adapt mechanism:}
According to Section \ref{sec:AttentiontoAdaptMechanism}, we introduce an attention-to-adapt mechanism. To better evaluate the performance of it, other possible structures are investigated as comparisons. 

\textit{Source-target subnet.} 
In order to evaluate the efficacy of this subnet, two additional structures are explored (Table \ref{tab:ATA}). In the first structure, no souce-target {subnet} is considered, and only source and target subnets are used for training (Simple-DA). This variant increases $E_{MAE}$ by 14.4 and reduces the $E_{F1}$ by 14.6\% across the three mentioned scenarios compared to AdaTreeFormer. In the second structure, a bidirectional source-target and target-source scheme are employed in the source-target subnet (Bi-DA). Using a bidirectional cross-domain attention, the results are improved compared to Simple-DA, but the performance is lower than that of AdaTreeFormer which has a unidirectional cross-domain attention (see Section \ref{sec:AttentiontoAdaptMechanism}). Since only a limited number of training images are available in the target domain, it is not appropriate to use a bidirectional structure for cross-domain attention. In fact, more attention should be paid to the target domain during training as does AdaTreeFormer.

\begin{table*}[!t]
\centering
	\caption{Ablation study of the attention-to-adapt mechanism on the $S_L \rightarrow S_J$, $S_J \rightarrow S_L$, and $S_Y \rightarrow S_L$ domain adaptation scenarios.}
	\begin{center}
	\begin{tabular}{c|c c |c c | c c}
    \toprule
	\midrule
		Scenario & \multicolumn{2}{c}{$S_L \rightarrow S_J$} & \multicolumn{2}{c}{$S_J \rightarrow S_L$} &  \multicolumn{2}{c}{$S_Y \rightarrow S_L$} \\
    \midrule
		Method  & $E_{MAE}$ & $E_{F1}$ & $E_{MAE}$ & $E_{F1}$ & $E_{MAE}$ & $E_{F1}$ \\
    \midrule
	  {Simple-DA} & 148.4 & 40.8 &  29.9 & 57.8 & 31.9 & 53.5\\  
        {Bi-DA} & 132.9 & 44.6 & 24.7 & 58.1 & 25.7 & 55.3 \\  
        {AdaTreeFormer}  & \textbf{122.5} & \textbf{53.8} & \textbf{21.4} & \textbf{71.5} & \textbf{23.1} & \textbf{70.8} \\         
	\bottomrule
	\end{tabular}
	\end{center}
    \label{tab:ATA}	
\end{table*}

\textit{Hierarchical domain attention.}
We employ the introduced domain attention block (DAB) in a hierarchical way at three scales (\ie $\frac{1}{8}, \frac{1}{16}$ and $\frac{1}{32}$) of the proposed framework (Section \ref{sec:AttentiontoAdaptMechanism}). To empirically assess the influence of hierarchical domain attention, we perform experiments to analyze the impact of applying the proposed DAB at varying scales. In this regard, when the DAB is applied only at the coarse scale of the decoder ($\frac{1}{32}$), it leads to an increase of 28.1, 10.8, and 14.1 in $E_{MAE}$ and a decrease of 12.6\%, 24.4\%, 28.2\% in $E_{F1}$ for the $S_L \rightarrow S_J$, $S_J \rightarrow S_L$, and $S_Y \rightarrow S_L$ scenarios (Table \ref{tab:DAB}). This decrease in accuracy and increase in error is less when DAB is applied in the $\frac{1}{8}$ and $\frac{1}{4}$ scales. However, the accuracy achieved by applying the DAB at two scales is still lower than when applying it at all three scales of the decoder. For instance, by employing DAB at both $\frac{1}{32}$ and $\frac{1}{16}$ scales, the $E_{MAE}$ increases by 8.3, 2.9, and 4.0, while the $E_{F1}$ decreases by 7.6\%, 9.2\%, and 3.5\% in the mentioned scenarios, in comparison to AdaTreeFormer.

\begin{table*}[!t]
\centering
	\caption{Ablation study of the DAB on the $S_L \rightarrow S_J$, $S_J \rightarrow S_L$, and $S_Y \rightarrow S_L$ domain adaptation scenarios.}
	\begin{center}
	\begin{tabular}{c c c|c c |c c | c c}
    \toprule
	\midrule
		\multicolumn{3}{c}{Scale} & \multicolumn{2}{c}{$S_L \rightarrow S_J$} & \multicolumn{2}{c}{$S_J \rightarrow S_L$} &  \multicolumn{2}{c}{$S_Y \rightarrow S_L$} \\
    \midrule
		$\frac{1}{8}$ & $\frac{1}{16}$ & $\frac{1}{32}$ & $E_{MAE}$ & $E_{F1}$ & $E_{MAE}$ & $E_{F1}$ & $E_{MAE}$ & $E_{F1}$ \\
    \midrule
	  &  & \cmark  &  150.6 & 41.2 & 32.2 & 47.1 & 37.2 & 42.6\\
        & \cmark &  & 137.2 & 50.7 & 28.8 & 62.8 & 29.9 & 63.1 \\  
        \cmark & & & 142.5 & 47.3 & 27.4 & 60.4 & 28.3 & 64.9\\  
        \cmark &  & \cmark & 128.3 & 49.2 & 25.4 & 62.4 & 26.9 & 65.2\\  
        & \cmark & \cmark &  130.8 & 46.2 & 24.3 & 62.3 & 27.1 & 67.3\\ 
        \cmark & \cmark &  & 125.3 & 51.9 & 23.4 & 66.3 & 24.7 & 65.2\\  
        \cmark & \cmark & \cmark & \textbf{122.5} & \textbf{53.8} & \textbf{21.4} & \textbf{71.5} & \textbf{23.1} & \textbf{70.8} \\         
	\bottomrule
	\end{tabular}
	\end{center}
    \label{tab:DAB}	
\end{table*}

\newpage

\begin{figure*}[!t]
\centering
\includegraphics[width=6in]{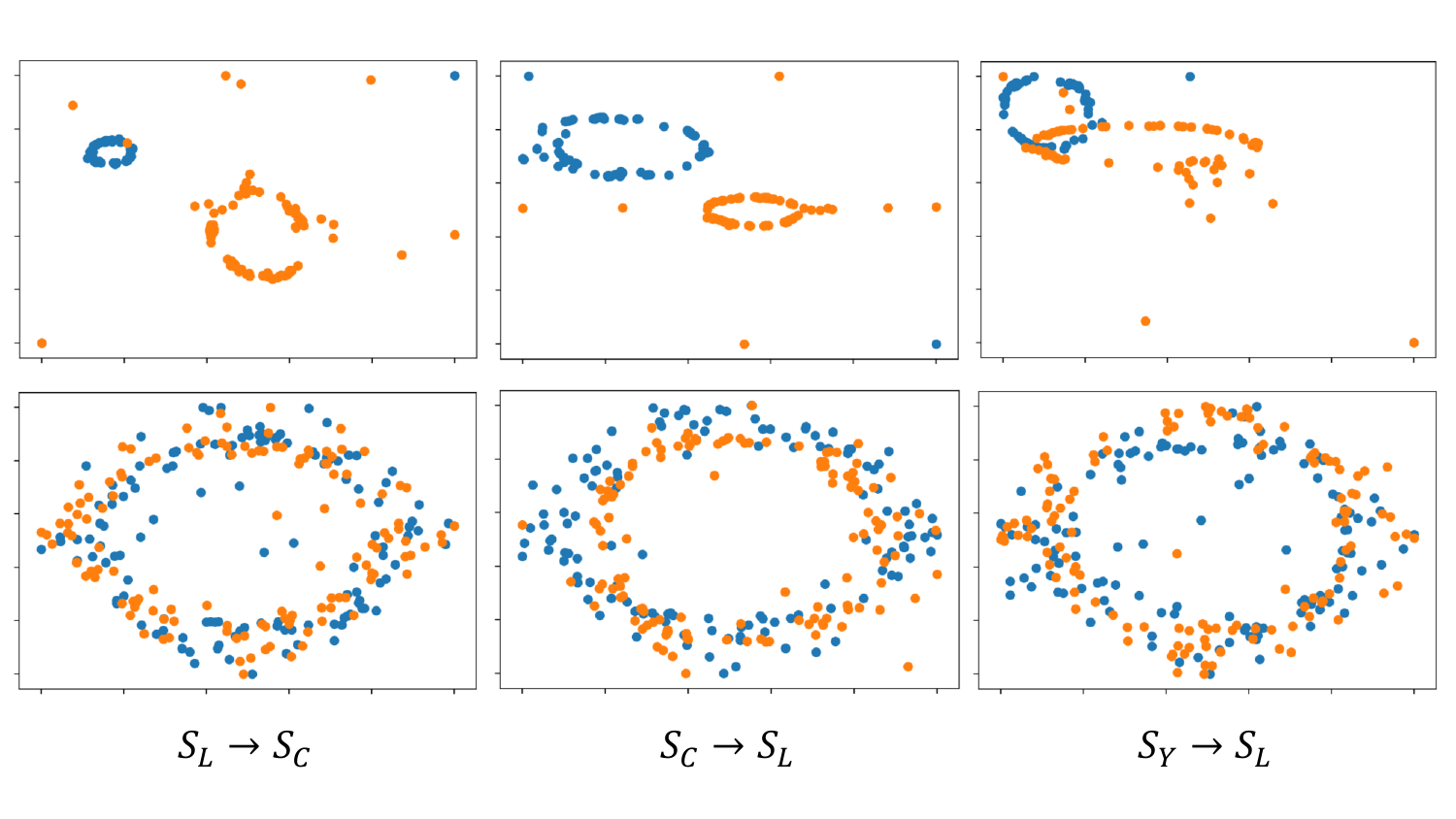}
\caption{t-SNE visualization of features extracted from source (blue) and target (orange) domain images. The top row represents the features processed through the HTFE without applying the attention-to-adapt mechanism. In contrast, the bottom row illustrates the features after applying the attention-to-adapt mechanism.}

\label{fig:tSNE}
\end{figure*}

\textit{t-SNE Visualization.} The primary objective of domain adaptation is to ensure that the feature distributions obtained from various domains are similar. In this section, we employed t-SNE to reduce the dimensionality of the feature map output through the feature extraction network, as depicted in Fig.~\ref{fig:tSNE}, to evaluate the effectiveness of the proposed domain adaptation method. The blue circles represent the outcomes obtained from the source domain, while the orange circles represent those obtained from the target domain in different tasks. It is evident that in the absence of domain adaptation, the model extracts feature distributions that are farther apart. This suggests that the model extracts distinct features from the source and target domains, making it challenging to produce accurate results. However, after implementing the proposed attention-to-adapt mechanism, the model can extract similar feature distributions from two different domains, enhancing its ability to obtain domain-invariant features from the target domain.

\begin{figure*}[!t]
\centering
\includegraphics[width=6in]{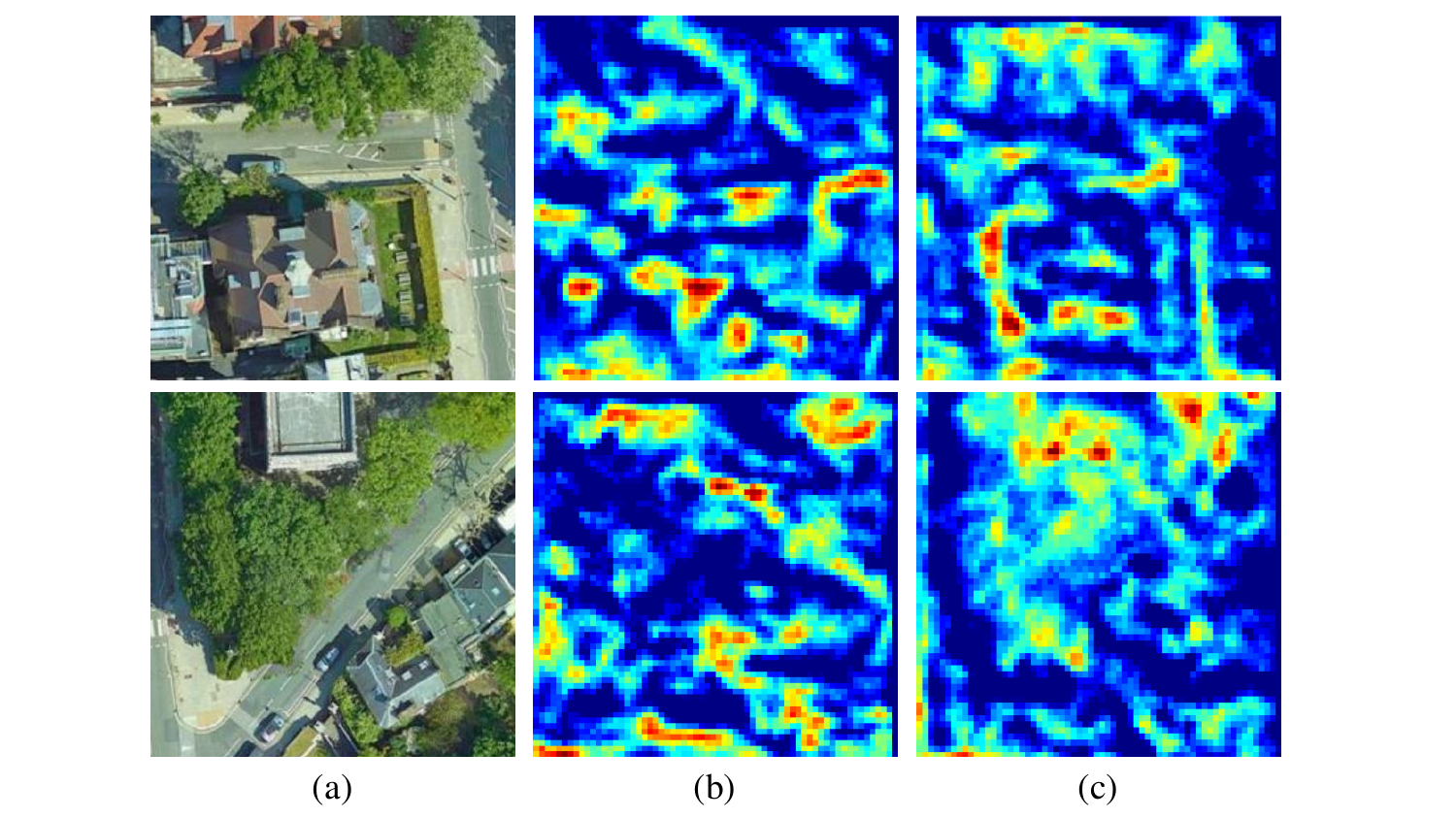}
\caption{(a) Sample images from the London dataset, the generated feature maps (b) without and (c) with the proposed domain adaptation technique. Notice only 5 shots from the London dataset are used as training images in the target domain.}

\label{fig:features_da}
\end{figure*}

In Fig.~\ref{fig:features_da}, we further offer two examples of the generated feature maps from the $A_{attn}^{'}$ of the model (Fig.~\ref{fig:overview}b) with and without the proposed domain adaptation technique in the $S_J \rightarrow S_L$ scenario. As one can see, without adaptation, the feature maps highlight various domain-specific objects such as roads, buildings, etc (Fig.~\ref{fig:features_da}b). This demonstrates the confusion of the model in extracting suitable tree related features in the target domain. While the generated feature maps with domain adaptation focus on the tree covered area (Fig.~\ref{fig:features_da}c), showing the robustness of our model in extracting suitable domain invariant features.

\subsubsection{Analysis on training loss}
In this section, we investigate the effectiveness of each component of the employed training loss. {In order to verify the effectiveness of the training loss, firstly, we use the $L2$ loss instead of the $L_{TDM}$ loss (w/ $L2$).} Table \ref{tab:TL} shows that using $L2$ increases the $E_{MAE}$ by 78.9, 13.9, and 15.2 for $S_L \rightarrow S_J$, $S_J \rightarrow S_L$, and $S_Y \rightarrow S_L$, respectively. 
Also, the $E_{F1}$ exhibits a reduction of 25.4\%, 48.1\%, and 49.5\%, respectively. The $E_{F1}$ evaluates a model's precision and recall in correctly identifying the location of objects within an image, providing a comprehensive measure of its localization performance.

Next, we analyze the performance of the proposed AdaTreeFormer without hierarchical cross domain feature alignment ($L_{HCDFA}$), as discussed in Section \ref{sec:HierarchicalCrossDomainFeatureAlignment}, denoted as ``w/o $L_{HCDFA}$", which increases $E_{MAE}$ by 2.5 and decreases $E_{F1}$ by 2.7\%.

Also, computing the $L_{HCDFA}$ only on the third scale of the decoder (Single  $L_{HCDFA}$) instead of three scales of the decoder (Fig.~\ref{fig:overview}) decreases the accuracy compared to the AdaTreeFormer. 
At last, the performance of the AdaTreeFormer without using adversarial learning is assessed (w/o $L_{Adv}$). Not using the adversarial learning averagely increases the $E_{MAE}$ by 2.6 and decreases $E_{F1}$ by 2.9\%.

\begin{table*}[!t]
\centering
	\caption{Ablation study of the employed training loss on the $S_L \rightarrow S_J$, $S_J \rightarrow S_L$, and $S_Y \rightarrow S_L$ domain adaptation scenarios.}
	\begin{center}
	\begin{tabular}{c|c c |c c | c c}
    \toprule
	\midrule
		Scenario & \multicolumn{2}{c}{$S_L \rightarrow S_J$} & \multicolumn{2}{c}{$S_J \rightarrow S_L$} &  \multicolumn{2}{c}{$S_Y \rightarrow S_L$} \\
    \midrule
		Method  & $E_{MAE}$ & $E_{F1}$ & $E_{MAE}$ & $E_{F1}$ & $E_{MAE}$ & $E_{F1}$ \\
    \midrule
	  {w/ $L2$} & 201.4 & 28.4 & 35.3 & 23.4 &  38.3 & 21.3\\  
        {w/o $L_{HCDFA}$} &  126.5 & 50.4 & 22.8 & 69.4 & 25.3 & 68.1 \\  
        {Single  $L_{HCDFA}$} & 124.3 & 52.2 & 22.0 & 70.3 & 24.7 & 69.3 \\  
        {w/o $L_{Adv}$} &  125.8 & 50.3 & 23.5 & 70.2 & 26.3 & 66.9 \\  
        {AdaTreeFormer} & \textbf{122.5} & \textbf{53.8} & \textbf{21.4} & \textbf{71.5} & \textbf{23.1} & \textbf{70.8}  \\         
	\bottomrule
	\end{tabular}
	\end{center}
    \label{tab:TL}	
\end{table*}

\newpage
We specifically investigate the components of $L_{HCDFA}$. In this regard, the impact of different weight values, denoted as $\beta_1$ and $\beta_2$, in the $L_{HCDFA}$ loss is investigated. These weight values determine the degree of similarity between the features extracted from the middle subnet and those obtained from the source and target domains. To this end, we change the weight values of $\beta_1$ and $\beta_2$ from 0.1 to 0.9 with an interval of 0.2 and assess the achieved results (Table \ref{tab:HCDFA}). According to the results, the network achieves the most accurate results when $\beta_1 =  0.3$ and $\beta_2 = 0.7$.

\begin{table*}[!t]
\centering
	\caption{Ablation study of the HCDFA loss on the $S_L \rightarrow S_J$, $S_J \rightarrow S_L$, and $S_Y \rightarrow S_L$ domain adaptation scenarios.}
	\begin{center}
	\begin{tabular}{c|c c |c c | c c}
    \toprule
	\midrule
		Scenario & \multicolumn{2}{c}{$S_L \rightarrow S_J$} & \multicolumn{2}{c}{$S_J \rightarrow S_L$} &  \multicolumn{2}{c}{$S_Y \rightarrow S_L$} \\
    \midrule
		Method  & $E_{MAE}$ & $E_{F1}$ & $E_{MAE}$ & $E_{F1}$ & $E_{MAE}$ & $E_{F1}$ \\
    \midrule
	  {$\beta_1 = 0.9$, $\beta_2 = 0.1$} & 140.5 & 50.0 & 26.7 & 60.2 & 28.4 & 59.1\\  
        {$\beta_1 = 0.7$,  $\beta_2 = 0.3$} & 132.8 & 51.2 & 24.4 & 63.0 & 26.1 & 63.1 \\  
        {$\beta_1 = 0.5$,  $\beta_2 = 0.5$} & 124.5 & 51.7 & 22.4 & 68.8 & 24.8 & 67.1 \\  
        {$\beta_1 = 0.3$,  $\beta_2 = 0.7$} & \textbf{122.5} & \textbf{53.8} & \textbf{21.4} & \textbf{71.5} & \textbf{23.1} & \textbf{70.8}  \\  
        {$\beta_1 = 0.1$,  $\beta_2 = 0.9$} &  123.8 & 53.2 & 22.0 & 71.4 & 25.3 & 68.5\\           
	\bottomrule
	\end{tabular}
	\end{center}
    \label{tab:HCDFA}	
\end{table*}

\begin{table*}[!t]
\centering
	\caption{Ablation study of using 1, 5, and 10-shot labeled data of target domain on the $S_L \rightarrow S_J$, $S_J \rightarrow S_L$, and $S_Y \rightarrow S_L$ domain adaptation scenarios.}
	\begin{center}
	\begin{tabular}{c|c c |c c | c c}
    \toprule
	\midrule
		Scenario & \multicolumn{2}{c}{$S_L \rightarrow S_J$} & \multicolumn{2}{c}{$S_J \rightarrow S_L$} &  \multicolumn{2}{c}{$S_Y \rightarrow S_L$} \\
    \midrule
		Method  & $E_{MAE}$ & $E_{F1}$ & $E_{MAE}$ & $E_{F1}$ & $E_{MAE}$ & $E_{F1}$ \\
    \midrule
	  {w/o } & 273.0  &  5.2 &  124.7 & 36.6  & 122.6  &  15.5 \\  
	  {1 shot} &  180.2 & 19.3  & 56.2  & 53.1  & 61.9 & 48.3  \\  
    {5 shot} & 122.5 & 53.8 & 21.4 & 71.5 & 23.1 & 70.8  \\    
    {10 shot} & \textbf{88.7} & \textbf{60.2} & \textbf{20.1} & \textbf{72.4} & \textbf{22.1} & \textbf{71.6}  \\  
	\bottomrule
	\end{tabular}
	\end{center}
    \label{tab:VarousShot}	
\end{table*}

\subsubsection{Analysis on selecting few-shot data}
 Given that our domain adaptation method necessitates a limited number of labeled images from the target domain, this section will explore the implications of choosing a varying number of few-shot data for AdaTreeFormer. In this regard, 1, 5, and 10 labeled images of the target domain are utilized for training (1, 5, and 10-shot learning) and the results are evaluated. According to Table \ref{tab:VarousShot}, using 1-shot learning compared to that without adaptation (w/o adaptation) causes decrease in the $E_{MAE}$ by 92.8 and 68.5 and 60.7 and increase in the $E_{F1}$ by 14.1, 16.5 and 32.8, for $S_L \rightarrow S_J$, $S_J \rightarrow S_L$, and $S_Y \rightarrow S_L$, respectively. As the number of employed labeled images from the target domain is raised from 1 to 5 and 10, there is a consistent increase in result accuracy. Employing 10-shot learning instead of 5-shot learning decreases the $E_{MAE}$ by 33.8, 1.3, and 1.0, and increases the $E_{F1}$ by 6.4, 0.9, and 0.8 for the $S_L \rightarrow S_J$, $S_J \rightarrow S_L$, and $S_Y \rightarrow S_L$, respectively.

\section{Conclusion} 
In this paper, we introduce a transform-based end-to-end few-shot domain adaptation framework for tree counting from a single high-resolution remote sensing image. In this network, an attention-to-adapt mechanism is introduced to produce robust self- and cross-domain attention maps for tree density estimation using features extracted from the source and target domains. In addition, we propose a hierarchical cross-domain feature alignment loss to guide the network in extracting robust features from the limited images of the target domain.  
Adversarial learning is adopted to force the network to reconcile the distribution of source and target domains. The attention-to-adapt mechanism significantly reduces the average absolute counting error compared to the finetuned baseline (\eg by 7.5 points in the cross domain from the London to Jiangsu dataset); subsequent inclusion of the hierarchical cross-domain feature alignment loss and adversarial learning further reduces the average absolute counting error (\eg by 4.0 points in London $\rightarrow$ Jiangsu).
The overall results in all designed domain adaptation scenarios indicate that our method outperforms the state of the art by large margins.

The core functionalities developed here, particularly the domain adaptation techniques, have the potential to be applied to a wider range of tasks beyond tree counting. For example, the framework could be adapted for counting other objects of urban significance such as buildings and vehicles, or ecological significance such as wildlife populations, in environments with varying visual characteristics. Furthermore, the domain adaptation strategies could be generalized to other object detection tasks in computer vision, such as crowd counting, where data from different domains is available. By demonstrating the effectiveness of cross-domain adaptation in tree counting, we pave the way for the development of more versatile and robust object counting models in diverse real-world applications.

\section*{Acknowledgements}
This project (ReSET) has received funding from the European Union’s Horizon 2020 FET Proactive Programme under grant agreement No 101017857. The contents of this publication are the sole responsibility of the ReSET consortium and do not necessarily reflect the opinion of the European Union. Miaojing Shi was supported by theFundamental Research Funds for the Central Universities.

\bibliographystyle{elsarticle-harv} 
\bibliography{Citation}

\end{document}